\documentclass{article}

\usepackage{microtype}
\usepackage{graphicx}
\usepackage{booktabs} 

\usepackage{hyperref}

 \usepackage[accepted]{icml2024}

\usepackage{amsmath}
\usepackage{amssymb}
\usepackage{mathtools}
\usepackage{amsthm}

\usepackage[capitalize,noabbrev]{cleveref}

\usepackage[utf8]{inputenc} 
\usepackage[T1]{fontenc}    
\usepackage{url}            
\usepackage{booktabs}       
\usepackage{amsfonts}       
\usepackage{nicefrac}       
\usepackage{microtype}      
\usepackage{lipsum}
\usepackage{bbm}
\usepackage{enumitem}
\usepackage{optidef}
\usepackage{subcaption}
\usepackage{bm}
\usepackage{upgreek}
\usepackage{multicol}
\usepackage{wrapfig}
\usepackage{setspace}
\usepackage{multirow}

\theoremstyle{plain}
\newtheorem{theorem}{Theorem}[section]
\newtheorem{proposition}[theorem]{Proposition}

\theoremstyle{definition}
\newtheorem{definition}[theorem]{Definition}

\theoremstyle{definition}
\newtheorem{example}[theorem]{Example}
\theoremstyle{remark}

\usepackage{thmtools,thm-restate}

\declaretheorem{fact}

\usepackage[textsize=tiny]{todonotes}

\def\ddefloop#1{\ifx\ddefloop#1\else\ddef{#1}\expandafter\ddefloop\fi}
\def\ddef#1{\expandafter\def\csname bb#1\endcsname{\ensuremath{\mathbb{#1}}}}
\ddefloop ABCDEFGHIJKLMNOPQRSTUVWXYZ\ddefloop
\def\ddef#1{\expandafter\def\csname c#1\endcsname{\ensuremath{\mathcal{#1}}}}
\ddefloop ABCDEFGHIJKLMNOPQRSTUVWXYZ\ddefloop
\def\ddef#1{\expandafter\def\csname h#1\endcsname{\ensuremath{\widehat{#1}}}}
\ddefloop ABCDEFGHIJKLMNOPQRSTUVWXYZ\ddefloop
\def\ddef#1{\expandafter\def\csname v#1\endcsname{\ensuremath{\boldsymbol{#1}}}}
\ddefloop ABCDEFGHIJKLMNOPQRSTUVWXYZabcdefghijklmnopqrstuvwxyz\ddefloop
\def\ddef#1{\expandafter\def\csname v#1\endcsname{\ensuremath{\boldsymbol{\csname #1\endcsname}}}}
\ddefloop {alpha}{beta}{gamma}{delta}{epsilon}{varepsilon}{zeta}{eta}{theta}{vartheta}{iota}{kappa}{lambda}{mu}{nu}{xi}{pi}{varpi}{rho}{varrho}{sigma}{varsigma}{tau}{upsilon}{phi}{varphi}{chi}{psi}{omega}{Gamma}{Delta}{Theta}{Lambda}{Xi}{Pi}{Sigma}{varSigma}{Upsilon}{Phi}{Psi}{Omega}{ell}\ddefloop

\newcommand{\Parents}{Pa}

\DeclareSymbolFont{symbolsC}{U}{txsyc}{m}{n}
\DeclareMathSymbol{\boxright}{\mathrel}{symbolsC}{128}

\def\*#1{\mathbf{#1}}

\newcommand{\Pai}[1]{\*{Pa}_{#1}}

\newcommand{\Ui}[1]{\*{U}_{#1}}

\DeclareMathOperator{\unif}{Unif}
\definecolor{myRed}{rgb}{0.855,0.23,0.15}
\definecolor{myGreen}{rgb}{0.505,0.84,0.325}
\definecolor{GGreen}{rgb}{0.305,0.7,0.225}

\DeclareMathOperator{\bern}{Bernoulli}

\usepackage{tikz}
\usetikzlibrary{shapes,decorations,arrows,calc,arrows.meta,fit,positioning}
\tikzset{
    -Latex,auto,node distance =1 cm and 1 cm,semithick,
    state/.style ={ellipse, draw, minimum width = 0.7 cm},
    point/.style = {circle, draw, inner sep=0.04cm,fill,node contents={}},
    bidirected/.style={Latex-Latex,dashed},
    el/.style = {inner sep=2pt, align=left, sloped}
}

\tikzstyle{SCM}=[>={Latex[length=3mm, width=1.5mm]},
    every node/.style={inner sep=0.25em, transform shape},
    every path/.append style={->,draw=black, transform shape}, 
    conf-path/.style={dashed, bend left=30},
    bd/.style={dashed, bend left=40, draw=black, every edge/.append style={<->}},
    neg/.style={every edge/.append style={draw=myRed}},
    pos-bd/.style={dashed, bend left=40, every edge/.append style={<->, draw=myGreen}},
    neg-bd/.style={dashed, bend left=40, every edge/.append style={<->, draw=myRed}},
    posi/.style={every edge/.append style={draw=myGreen}},
    removed/.style={opacity=0.2,every label/.style={opacity=0.2}},
    cut/.style={color=Red!50},
    sel-node/.style={rectangle,inner sep=0.3em, draw, color=MidnightBlue, fill=MidnightBlue,node contents={}},
    sig-node/.style={rectangle,inner sep=0.3em, draw, color=Green,node contents={}},
    s-node/.style={double distance=0.5mm, circle,draw},
    point/.style = {circle, draw, inner sep=0.06cm,fill,node contents={}},
    l-point/.style = {circle, draw, inner sep=0.06cm, node contents={}},
    t-node/.style={rectangle,inner sep=0.3em, draw,node contents={}},
    hl1/.style={fill=blue,color=blue,line width=0.7mm},
    hl2/.style={fill=red,color=red,line width=0.7mm},
    redbox/.style={fill=red!80,opacity=0.2,color=red,draw=red},
    bluebox/.style={fill=blue!80,opacity=0.2,color=blue,draw=blue},
    greenbox/.style={fill=green!80,opacity=0.2,color=green,draw=green},
    cond/.style={anchor=north,yshift=0.3em,shape=rectangle, draw, minimum width=1.1em, minimum height=2.2em, blue, thick},
    flow/.style={-,decorate, draw=blue, decoration={snake=wave,amplitude=.7mm},thick}
]

\icmltitlerunning{Counterfactual Image Editing}

\begin{document}

\twocolumn[
\icmltitle{Counterfactual Image Editing}



\icmlsetsymbol{equal}{*}

\begin{icmlauthorlist}
\icmlauthor{Yushu Pan}{yyy}
\icmlauthor{Elias Bareinboim}{yyy}
\end{icmlauthorlist}

\icmlaffiliation{yyy}{Department of Computer Science, Columbia University, New York, USA}

\icmlcorrespondingauthor{Yushu Pan}{yushupan@cs.columbia.edu}
\icmlcorrespondingauthor{Elias Bareinboim}{eb@cs.columbia.edu}

\icmlkeywords{Machine Learning, ICML}

\vskip 0.3in
]



\printAffiliationsAndNotice{}  

\begin{abstract}
Counterfactual image editing is an important task in generative AI, which asks how an image would look if certain features were different.  
The current literature on the topic focuses primarily on changing individual features while remaining silent about the causal relationships between these features, as present in the real world. 
In this paper, we formalize the counterfactual image editing task using formal language, modeling the causal relationships between latent generative factors and images through a special type of model called augmented structural causal models (ASCMs). 
Second, we show two fundamental impossibility results: (1) counterfactual editing is impossible from i.i.d. image samples and their corresponding labels alone; (2) even when the causal relationships between the latent generative factors and images are available, no guarantees regarding the output of the model can be provided.
Third, we propose a relaxation for this challenging problem by approximating non-identifiable counterfactual distributions with a new family of \textbf{counterfactual-consistent estimators}.  
This family exhibits the desirable property of preserving features that the user cares about across both factual and counterfactual worlds. 
Finally, we develop an efficient algorithm to generate counterfactual images by leveraging neural causal models. 
\end{abstract}

\begin{figure*}[t]
    \centering
    \includegraphics[scale=1.3]{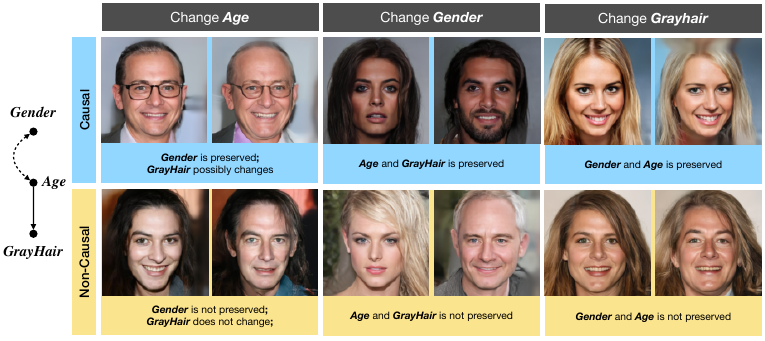}
    \caption{(Left) A causal graph depicting the causal relationships among features. (Right) Image editing results are displayed, with the first row showing edits incorporating causal relations, and the second row without them. 
    Each column represents a unique counterfactual query, altering the age, gender, and gray hair of the individuals. 
    These instances provide preliminary evidence that the causal approach introduced in this paper ensures the preservation of the relevant causal invariances for the query across both factual and counterfactual images. 
    }
\vspace{-9pt}
\label{fig:face-ex}
\end{figure*}

\vspace{-0.2in}
\section{Introduction}
\vspace{-0.05in}

Counterfactual reasoning is a critical component of our cognitive system. It is essential for solving various tasks, including assigning credit, determining blame and responsibility, understanding why events occurred in a particular way and articulating explanations, and generalizing across changing conditions and environments \citep{pearl:mackenzie2018,bareinboim:etal20,correa2021nested}.
More recently, there has been a growing interest in counterfactual questions regarding image generation and editing. 
For instance, one might ask ``how would the image change had the dog been a cat?'' or ``What would the image look like had the person been smiling?''.
Addressing these prototypical counterfactual questions is challenging and requires the understanding of the causal relationships between the features, with practical applications in various downstream tasks, including data augmentation, fairness analysis, generalizability, and transportability ~\cite{bareinboim:etal15, scholkopf2021toward, lee2020generalized, mao2022causal}.

Some initial methods for counterfactual image editing tasks typically involve searching for adversarial samples ~\cite{goyal2019counterfactual, wang2020scout, dhurandhar2018explanations}. For example, ~\cite{dhurandhar2018explanations} proposed a minimum-edit counterfactual method that aims to identify the minimum and most effective perturbations needed to change the classifier's prediction. With the ability to generate high-quality synthetic images from a latent space through GANs ~\cite{brock2018large, karras2019style}, VAEs ~\cite{child2021very, vahdat2020nvae}, and Diffusion Models ~\cite{ho2020denoising, song2020score}, recent approaches edit images by manipulating vectors in the latent space ~\cite{shen2020interpreting, harkonen2020ganspace, khorram2022cycle, chai2021using}. 

More recently, text information has also been leveraged in image editing tasks. The image description in text is beneficial to the encoding process and guiding manipulations in the latent space \citep{radford2021learning, avrahami2022blended, crowson2022vqgan, gal2022stylegan, patashnik2021styleclip} and the natural editing instruction text can be directly used to prompt the transition from the original tothe counterfactual images \citep{brooks2023instructpix2pix}. 
However, such approaches focus primarily on changing a single categorical label of a given image, and more fundamentally, do not take the causal relationships among the underlying generative factors into account. The next example illustrates the challenge when multiple features are involved in the generation.
\begin{example}
\label{ex:face}
Consider an image dataset of human faces. 
Based on our understanding of human anatomy and facial expressions, we know that both  $Gender$ and $Age$ do not causally affect each other while age does affect hair color. Meanwhile, the dataset collected has older males and younger females, i.e., there exists a strong correlation between age and gender. 
Formally, the causal relationships between these three generative factors are shown in Fig.~\ref{fig:face-ex}. 

Existing methods focus on the editing of a single concept while the effects of the intervened concepts on others are not taken into account. 
Suppose we are evaluating the counterfactual query: "Given a certain image, what would the face look like had she been older?".  
If the age of the person is changed naively, gender and hair color may also change due to the correlation between these features found in the data. 
For example, when making an image of a woman older, it may inadvertently also change her gender to male; see the yellow row in \cref{fig:face-ex}. However, 
it would be expected that changes in age should not affect gender when performing causal editing, as shown in the figure's first row (in blue).

More importantly, existing methods are unable to answer to \textbf{what extent hair color should change after an intervention on age.}
Even though some recent methods may be able to enforce consistency in terms of gender, the causal effect from the age to the hair color may not be reflected correctly in the counterfactual images. 
For instance, gray hair may never appear after the editing by non-causal approaches. 
In contrast, causal image editing ensures the effects of target interventions on other features are carried over properly from factual to the proper counterfactual world. 
To illustrate, edits in Fig.~\ref{fig:face-ex} (blue) are more closely aligned with the reality in which these causal invariances are presented.
\hfill $\blacksquare$
\end{example}

To capture the causal relationships among generative factors, we build on a class of generative models known as Structural Causal Models (SCMs) \citep{pearl:2k}. A fully instantiated SCM induces what is known as the Pearl Causal Hierarchy (PCH; also called \textit{ladder of causation}) \citep{pearl:mackenzie2018,bareinboim:etal20}. 
The PCH consists of families of distributions in increasing levels of refinement: layer 1  ($\cL_1$) corresponds to passive observations and typical correlations, layer 2 ($\cL_1$) to interventions  (e.g., changing a variable to see the effect), and layer 3 ($\cL_3$) to counterfactuals (e.g., considering what would happen under hypothetical scenarios). 
A result known as the causal hierarchy theorem states that higher-layer distributions cannot be answered only from the lower-layer ones \citep{bareinboim:etal20}. 

Recently, researchers have connected SCMs with deep generative models by implicitly finding surrogate models of the true generative model relating images and its generative factors. 
Despite the progress made so far, many of these works have limitations in different dimensions important in our context. 
First, they assume Markovianity, which implies the absence of unobserved confounding among generative factors. While this assumption may hold in specific settings, the same is certainly strong and does not hold in many others \citep{kocaoglu2018causalgan, pawlowski2020deep, sanchez2022diffusion, sauer2021counterfactual}. 

Second, many of these works estimate counterfactual queries for images and generate samples without considering whether the target query is identifiable. In particular, samples are generated even though the query is non-identifiable, which implies that no guarantee can be provided in terms of the quality and causal consistency of the image. In particular, it is unclear whether the causal invariances present in the real systems are preserved across the original and generated images. 

Third, other works focus on parametric SCMs over generative factors, such as linear mechanisms, while we study a more general class of non-parametric models \cite{yang2021causalvae, shen2022weakly}. 
Recently, a new class of generative models has been developed, the Neural Causal Model (NCM), which encodes causal constraints into deep generative models. These models are capable of both identifying and then estimating counterfactual quantities in non-parametric settings \citep{xia:etal21, xia:etal22}. 
Despite the soundness of this approach to handling general, non-parametric variables in theory, it remains challenging to estimate counterfactual images, as the structure between generative factors and images is not taken into account and it's hard to scale these models to higher dimensions. Further discussion on related works is provided in \cref{sec:relatedwork}.

In this paper, we study the principles underpinning counterfactual image editing tasks and develop a practical, causally-grounded framework for these critical generative capabilities for high-dimensional settings. 
To achieve this goal, we formalize counterfactual image tasks according to augmented SCMs (ASCMs), a special class of SCMs taking the image generation step into account.
This formulation allows for the formal encoding of causal relationships between generative factors and the image. It also enables modeling of the image editing tasks as querying counterfactual distributions induced by the true yet unknown ASCMs. 
More specifically, our contributions are as follows:

1. We formally show that image counterfactual distributions are almost never identifiable from only observational i.i.d image samples. Further, even when the causal relationships between generative factors and images are given, the target counterfactual distribution is still non-identifiable (Sec.~\ref{sec:id}).

2. We relax these settings and develop a new family of \textbf{counterfactual (Ctf-) consistent estimators} to approximate non-identifiable distributions. 
This provides the first procedure with formal guarantees of causal consistency w.r.t. the true generative model. 
With a sufficient condition to obtain Ctf-consistent estimators, 
we then develop an efficient algorithm (ANCMs) to sample counterfactual images in practice (Sec.~\ref{sec:part-est}). 
Extensive experiments are conducted to demonstrate the effectiveness of ANCMs (Sec.~\ref{sec:exp}).

\subsection{Preliminary}
In this section, we provide the necessary background to understand this work. An uppercase letter $X$ indicates a random variable and a lowercase letter $x$ indicates
its corresponding value; 
bold uppercase $\*X$ denotes a set of random variables, and lowercase letter $\*x$ is its corresponding values. 
We use $\mathcal{X}_{X}$ to denote the domain of $X$ and $\mathcal{X}_{\*X} = \mathcal{X}_{{X}_1} \times \cdots \times \mathcal{X}_{{X}_d}$ for $\*X = \{X_1, \dots, X_d\}$. We denote $P(\*X)$ as a probability distribution over a set of random variables $\*{X}$ and $P(\*X=\*x)$ as the probability of $\*X$ being equal to the value of $\*x$ under the distribution $P(\*X)$.

Our work relies on the basic semantical framework structural causal models (SCMs) \citep[Ch.~7]{pearl:2k}; we follow the presentation in \citep{bareinboim:etal20}. 
\begin{definition}[Structure Causal Model(SCM)]
A Structure Causal Model (for short, SCM) is a 4-tuple  $<\*{U}, \*{V}, \mathcal{F}, P({\*{U}})>$, where \\
    (1) $\*{U}$ is a set of background variables, also called exogenous variables, that are determined by factors outside the model;\\
    (2) $\*{V} = \{V_1, V_2, \dots, V_d\}$ is the set of endogenous variables that are determined by other variables in the model;\\
    (3) $\mathcal{F}$ is the set of functions $\{f_{V_1}, f_{V_2}\dots, f_{V_d}\} $ mapping $\*U_{V_j} \cup \Pai{V_j}$ to $V_j$, where $\*U_{V_j} \subseteq \*{U}$ and $\*{Pa}_{V_j} \subseteq \*V \backslash V_j$;\\
    (4) $P(\*{U})$ is a probability function over the domain of $\*{U}$. \hfill $\blacksquare$
\end{definition} 
Each SCM $\cM$ induces a causal diagram $\cG$, which is a directed acyclic graph where every $V_j$ is a vertex. There is a directed arrow from $V_j$ to $V_k$ if $V_j \in \Pai{V_k}$. And there is a bidirected arrow between $V_j$ and $V_k$ if $\*{U}_{V_j}$ and $\*{U}_{V_k}$ are not independent with each other \citep[Def. 11]{bareinboim:etal20}. 

An intervention on a subset of $\*X \subseteq \*V$, denoted by $do(\*x)$, is an operation where $\*X$ takes value $\*x$, regardless how $\*X$ are originally defined. For an SCM $\cM$, let $\cM_{\*x}$ be the submodel of $\cM$ induced by $do(\*x)$. For any subset $\*Y \subseteq \*V$, the potential outcome $\*Y_{\*x}(\*u)$ is defined as the solution of $\*Y$ after feeding $\*U = \*u$ into the submodel $\cM_{\*x}$. Then $\*Y_{\*x}$ is called a counterfactual variable induced by $\cM$. Specifically, the event $\*Y_{\*x} = \*y$ represent "$\*Y$ would be $\*y$ had $\*X$ been $\*x$". The counterfactual quantities induced by an SCM $\cM$ are defined as \citep[Def. 7]{bareinboim:etal20}:
\begingroup\abovedisplayskip=0.5em\belowdisplayskip=0pt
 \begin{align}
    \label{eq:def:l3-semantics-nested}
    P^{\cM}(\*y_{\*x}, \dots, \*z_{\*w}) = 
    \int_{\cX_{\mathbf{U}}} \mathbbm{1}_{\*Y_{{\*x}}(\*u)=\*y,\dots, \*Z_{{\*w}}(\*u) = \*z} dP(\*u),
\end{align}
\endgroup
where $\*Y, \dots, \*Z, \*X, \dots, \*W \subseteq \*V$. Specifically, $P(\*Y_{\*x})$ reduces to an observational distribution $P(\*Y) $ taking $\*X$ as an empty set.

Given the observed distribution $P(\*V)$ and causal diagram $\cG$, the optimal counterfactual bounds are closed intervals based on the following optimization problem \citep{zhang:bareinboim21b}. 
\begin{definition}[Optimal Counterfactual Bounds] 
\label{def:ctf-optimal-bound}
For a causal diagram $\cG$ and observed distributions $P(\*V)$, the \emph{optimal bound} $[l, r]$ over a counterfactual probability $P^{\cM}(\*y_{\*x}, \dots, \*z_{\*w})$ is 
defined as, respectively, the minimum and maximum of the following optimization problem:
\begin{align}
\label{eq:def:ctf-optimal-bound}
\underset{\cM \in \Omega(\cG)}{\max / \min} \ \  P^{\cM}(\*y_{\*x}, \dots, \*z_{\*w}) \ \  \mathrm{s.t. } P^{\cM}(\*V) = P(\*V)
\end{align}
where $\Omega(\cG)$ is the space of all SCMs that agree with the diagram $\cG$, i.e., $\Omega(\cG) = \{\forall \cM|\cG_{\cM} = \cG\}$. \hfill $\blacksquare$
\end{definition}
By the formulation of \cref{eq:def:l3-semantics-nested}, all possible values of counterfactual query induced by SCMs that agree with the observational distributions and causal diagram are contained in the closed interval $[l, r]$.

We use neural causal models (NCMs) for estimating counterfactual distributions, which are defined as follows \citep{xia:etal21}:
\begin{definition}[$\cG$-Constrained Neural Causal Model ($\cG$-NCM)]
    \label{def:gncm}
    Given a causal diagram $\cG$, a $\cG$-constrained Neural Causal Model (for short, $\cG$-NCM) $\widehat{M}(\bm{\theta})$ over variables $\*V$ with parameters $\bm{\theta} = \{\theta_{V_i} : V_i \in \*V\}$ is an SCM $\langle \widehat{\*U}, \*V, \widehat{\cF}, \widehat{P}(\widehat{\*U}) \rangle$ such that $\widehat{\*U} = \{\widehat{U}_{\*C} : \*C \subseteq \*V\}$, where\\
    (1) each $\widehat{U}$ is associated with some subset of variables $\*C \subseteq \*V$, and $\cD_{\widehat{U}} = [0, 1]$ for all $\widehat{U} \in \widehat{\*U}$; \\
    (2) $\widehat{\cF} = \{\hat{f}_{V_i} : V_i \in \*V\}$, where each $\hat{f}_{V_i}$ is a feed forward neural network parameterized by $\theta_{V_i} \in \bm{\theta}$ mapping values of $\Ui{V_i} \cup \Pai{V_i}$ to values of $V_i$ for $\Ui{V_i} = \{\widehat{U}_{\*C} : \widehat{U}_{\*C} \in \widehat{\*U} \text{ s.t. } V_i \in \*C\}$ and $\Pai{V_i} = \Parents_{\cG}(V_i)$; \\
    (3) $\widehat{P}(\widehat{\*U})$ is defined s.t.\ $\widehat{U} \sim \unif(0, 1)$ for each $\widehat{U} \in \widehat{\*U}$.     \hfill $\blacksquare$
\end{definition}

\section{Augmented SCMs and Image Counterfactual Distributions}
\label{sec:problem_statement}
In this section, we model the image counterfactual editing problems in causal language. We first define a special type of SCM to model the generation process of an image variable $\*I$, which is called the Augmented SCM (ASCM).

\begin{definition}[Augmented Structure Causal Model]
\label{def:ascm}
An Augmented Structure Causal Model (for short, ASCM) over a generative level SCM $\cM_{0} = \langle \{\*U_0, \*V_0, \cF_0, P^0(\*U_{0})\} \rangle$ is a tuple $\cM = \langle \*U, \{\*V, \*I\}, \cF, P(\*U) \rangle$ such that \\
(1) exogenous variables $\*U = \{\*U_{0}, \*U_{\*I}\}$; \\
(2) $\*V  = \*V_0$ are labeled observed endogenous variables; $\*I$ is an $m$ dimensional image variable; \\
(3) $\cF = \{\cF_{0}, f_{\*I}\}$,  
where $f_{\*I}$ maps from (the respective domains of) $\*V \cup {\*U}_{\*I}$ to $\*I$, which is an invertible function regarding $\*V$. Namely, there exists a function $h$ such that $\*V = h(\*I)$. \\
(4) $P(\*U_0) = P^0(\*U_0)$. \hfill $\blacksquare$
\end{definition}
 \begin{figure}[t]
 \centering
  \begin{tabular}{ccc||c}
        $F$ & $Y$ & $H$ & $P(F, Y, H)$  \\ \hline 
        0     & 0     & 0    & 0.216             \\ \hline
        0     & 0     & 1    & 0.144             \\ \hline
        0     & 1     & 0    & 0.128           \\ \hline
        0     & 1     & 1    & 0.032				\\ \hline
        1     & 0     & 0    & 0.144             \\ \hline
        1     & 0     & 1    & 0.096             \\ \hline
        1     & 1     & 0    & 0.192           \\ \hline
        1     & 1     & 1    & 0.048
	\end{tabular}
 \vspace{-5pt}
        \caption{$P(\*V)$ induced by the ASCM in Example. \ref{ex:face-ascm}.
        \label{tab:face} }  
\vspace{-10pt}
\end{figure} 
The ASCM $\cM$ is in fact a "larger" SCM describing a two-stage generative process, where first the low-dimensional generative factors are produced and second these generative factors are mapped to a high-dimensional image. More specifically, the 
$\*U_{\*I}$ interact with labeled $\*V$ to produce other unlabeled features $\tilde{\*U}$ through part of $f_{\*I}$ in the first stage.
In the second stage, the remaining part of $f_{\*I}$ mixes the observed $\*V$ and unobserved generative factors $\tilde{\*U}$ to create the image's set of pixels. Notice that $\tilde{U}$ is not a part of $\*U_{\*I}$ since $\tilde{U}$ can be generated from $\*V$ plus $\*U_{\*I}$.
Throughout this paper, we assume that domains of observed generative factors $\*V$ are discrete and finite. 
An important aspect of $f_{\*I}$ is that it is invertible regarding $\*V$ since these generative factors $\*V$ are present directly in a given image $\*i$. 
This assumption is commonly used in non-linear ICA and representation learning literature \citep{locatello2019challenging, lachapelle2021disentanglement, hyvarinen1999nonlinear, khemakhem2020variational}.
The inverse $h$ represents a labeling process that assigns the correct labels of $\*V$ to $\*i$. Then, for any $\*W \subseteq \*V$: 
\begin{equation}
\label{eq:bij}
P(\*w \mid \*i) = \left\{
\begin{aligned}
&1  & \*w=h_{\*W}(\*i)\\
&0  &otherwise,\\
\end{aligned}
\right.
\end{equation}
where $h_{\*W}(\cdot)$ is the subfunction of $h$ mapping from $\*I$ to $\*W$.
The next example illustrates the modeling of face images discussed earlier.
\begin{example}
\label{ex:face-ascm}
(\cref{ex:face} continued). Now we consider the augmented generative process, ASCM 
$\cM^* = \langle \*U=\{U_F, U_{Y}, U_{H_1}, U_{H_2}, \*U_{\*I}\}, \{\{F, H, Y\}, \*I\}, 
 \cF^*, P^*(\*U) \rangle$, where the mechanisms 
\begin{equation}
{\cF^*} = \left\{
\begin{aligned}
F & \leftarrow U_F \oplus U_Y \\
Y & \leftarrow U_Y \\
H & \leftarrow (\neg Y \land U_{H_1}) \oplus (Y \land U_{H_2})  \\
\*I & \leftarrow {f}_{\*I}^{\mathrm{face}}(F, Y, H, \*U_{\*I})
\end{aligned}
\right.
\end{equation}
and the exogenous variables $U_F, U_Y, U_{H_1}, U_{H_2}$ are independent binary variables, and $P(U_F = 1) = 0.4, P(U_{Y} = 1) = 0.4, P(U_{H_1} = 1) = 0.4, P(U_{H_2} = 1) = 0.2$. $\*U_{\*I}$ can be correlated with $U_F, U_Y, U_{H_1}, U_{H_2}$.

The variable $F$ represents gender (male $F=0$; female $F=1$), $Y$ represents age (young $Y=0$; old $Y=1$), and $H$ represents whether the person has gray hair (gray $H=1$; non-gray $H=0$). The observational distribution $P(F, Y, H)$ induced by $\cM^*$ is shown in \cref{tab:face}. 
From the distribution, we can see that $Y=1$ and $H=1$ are positively correlated ($P(Y=1, F=1) > P(Y=1, F=0), P(Y=0, F=1)< P(Y=0, F=0)$), and older people are more likely to have gray hair $P(H=1 \mid Y=0) > P(H=1 \mid Y=1)$. 

 \begin{figure*}[t!]
    \begin{subfigure}[b]{1\columnwidth}
    \centering
    \includegraphics[scale=1.35]{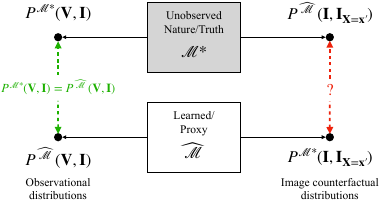}
    \caption{}
    \label{fig:image-collapse-l}
  \end{subfigure} \hfill
  \begin{subfigure}[b]{1\columnwidth}
        \centering
    \includegraphics[scale=1.3]{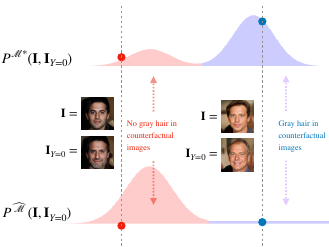}
    \caption{}
    \label{fig:image-collapse-r}
\vspace{-10pt}
  \end{subfigure} 
    \caption{ (a) The proxy generator $\widehat{\cM}$ is compatible with the same observational distributions with the unobserved true model but is not guaranteed to induce the same image counterfactual distributions. (b) Two different image counterfactual distributions in \cref{ex:face-collapse}. 
    Each sample from a $P(\*I, \*I_{Y=0})$ has an initial image $\*i$ and a counterfactual image $\*i'_{Y=0}$. Sampling from the red part of distributions, counterfactual images do not contain gray hair. Sampling from the blue part of distributions, counterfactual images have gray hair.
    }
\end{figure*}

Before the image is taken, $\*U_{\*I}$ and $\{F, Y, H\}$ produce other unobserved generative factors $\tilde{\*U}$, such as wrinkles, smiling, and narrow eyes at the generative level. Among them, some factors (such as wrinkles) can be produced by both $\*V$ and $\*U_{\*I}$, and some other factors (such as smiling) can be only produced by $\*U_{\*I}$. Then, $f_{\*I}$ maps all generative factors (including unobserved and observed ones) to image pixels $\*I$ at the second stage. Looking at the image, $\{F, Y, H\}$ are deterministic and one can in principle label them through function $h$, the inverse of ${f}_{\*I}^{\mathrm{face}}$ $\{F, Y, H\}$. 
\hfill $\blacksquare$
\end{example}

Equipped with ASCMs, we now formalize the counterfactual image generation tasks through formal causal semantics. Suppose the true underlying ASCM is given by $\cM^*$, which is unobserved. The goal is to query a specific type of counterfactual distribution induced by $\cM^*$ given the input distribution $P(\*V, \*I)$, i.e., $P^{\cM^*}(\*I, \*I_{{\*x'}})$, where $\*X \subseteq \*V$. Factorizing this joint probability distribution, we can write: 
\begin{equation}
\label{eq:img-ctf-query-decomp}
\begin{split}
    & P^{\cM^*}(\*I = \*i, \*I_{{\*x'}} = \*i') \\
    & = P^{\cM^*}(\*I = \*i)P^{\cM^*}(\*I_{{\*x'}} = \*i' \mid \*I = \*i ),
\end{split}
\end{equation}
this $\cL_3$-quantity can be explained as follows. The initial image $\*i$ is sampled from $P^{\cM^*}(\*I)$ and the goal is to edit $\*i$ to a counterfactual version $\*i'$ with modified features $\*X = \*x'$, where $\*i'$ is sampled from $P^{\cM^*}(\*I_{{\*x'}} \mid \*I = \*i )$ \footnote{$P^{\cM^*}( \*I_{{\*x'}} \mid \*I = \*i)$ serves for editing real images when the initial image $\*i$ is a real one given by an user.}. For example, the distribution $P^{\cM^*}(\*I, \*I_{Y=0})$ (induced by the ASCM introduced in \cref{ex:face-ascm}) can answer the query "generate an image describing people's face and edit the face to make the person look older".


 Throughout this paper, we call this type of $\cL_3$-distributions as \textit{Image Counterfactual Distributions}. A particular instantiation of the image variable, such as $P(\*I = \*i, \*I_{\*x'} = \*i')$, is called on \textit{Image Counterfactual Query}. The explanation of image counterfactual distributions at the generative level is that given all generative factors in the initial images, what would they be had $\*X$ taken value $\*x'$. For instance, $P^{\cM^*}(\*I, \*I_{Y=0})$ is asking what would observed factors (gender, hair color) and unobserved factors (wrinkles, smiling, narrow eyes, ...) be had the person been older.

\section{Non-identifiability of Image counterfactual Distributions}
\label{sec:id}

In classic counterfactual image generation tasks, a generator $\widehat{\cM}$ is trained to match the distribution $P(\*V, \*I)$ (e.g., through a GAN), and then the pair of an initial image and its counterfactual can be sampled from $P^{\widehat{\cM}}(\*I, \*I_{{\*x'}})$ induced by the generator (see \cref{fig:image-collapse-l} bottom). For concreteness, after sampling an initial image $\*i$, one can get the counterfactual image $\*i_{\*x'}$ by manipulating the latent space or conditional signal of $\widehat{\cM}$.
However, as alluded to earlier, the Causal Hierarchy Theorem \citep[Thm. 1]{bareinboim:etal20} states that counterfactual distributions cannot be computed merely from correlations. In particular, we show next the (non-)identifiability of any image counterfactual query from pure observational data:
\begin{restatable}[Image Causal Hierarchy Theorem]{corollary}{imagecht}
    \label{thm:image-cht}
    Any image counterfactual distribution is almost never uniquely computable from the observational distribution (or its samples). 
    \hfill $\blacksquare$
\end{restatable}
In other words, Corol. \ref{thm:image-cht} states that $P^{\widehat{\cM}}(\*I, \*I_{{\*x'}})$ induced by the proxy generator may not be consistent with the true $P^{{\cM}^*}(\*I, \*I_{{\*x'}})$ even when the proxy generator fits the observed distributions perfectly (i.e., $P^{\widehat{\cM}}(\*V, \*I) = P^{{\cM}^*}(\*V, \*I)$). This inconsistency implies the effect of intervention $\*X=\*x'$ on other generative factors (features) may differ from the true model and the proxy generator.

To illustrate this issue, suppose the target query is $P^{\cM^*}(\*I, \*I_{Y=0})$, where $\cM^*$ is defined in \cref{ex:face-ascm}. As shown in \cref{fig:image-collapse-r} and the next example, even when $\widehat{\cM}$ matches the observational distribution of the true underlying model $\cM^*$, $\widehat{\cM}$ can be less likely to generate counterfactual images with gray hair than $\cM^*$ after the intervention $Y=0$. 

\begin{example}
\label{ex:face-collapse}
Consider an ASCM $\cM'$ that is exactly the same as $\cM^*$ introduced in \cref{ex:face-ascm} except that
\begin{equation}
    f'_{H} = (\neg U_Y \land U_{H_1}) \oplus (U_Y \land U_{H_2})
\end{equation}

$\cM'$ implies the same $P(\*V)$ as shown in \cref{tab:face}. 
Then, it's immediately verifiable that $P^{\cM^*}(\*V, \*I) = P^{\cM'}(\*V, \*I)$.

Now consider the counterfactual query $P(\*i , \*i_{Y=0})$, where $\*i$ is a young male without gray hair (with generated features $F=0, Y=1, H=0$) and $\*i'$ is an old male with gray hair (with features $F=0, Y=0, H=1$). In $\cM^*$, $H=0$ will change to $H=1$ and $F$ will remains invariant after the intervention $do(Y=1)$ with probability $0.4$, i.e.: 
\begin{equation}
\label{eq:face-doage-hair}
    P^{\cM^*}(F_{Y=0}=0, H_{Y=0}=1 \mid F=0, Y=1, H=0) = 0.4. 
\end{equation}
However, $H$ will never change after the same intervention in $\cM'$ since the input of $f'_{H}$ does not involve $Y$, i.e., 
\begin{equation}
P^{\cM'}(F_{Y=0}=0, H_{Y=0}=1 \mid F=0, Y=1, H=0) = 0.
\end{equation}
Consequently, the true model suggests that the hair color is likely to change after making a young male look older with probability $0.4$ (see blue part of the distribution in the upper part of \cref{fig:image-collapse-r}) while the counterfactual image generated by the proxy model would have gray hair with zero probability (blue part of upper distribution in \cref{fig:image-collapse-r}). This is an instance of the aforementioned non-identifiability result (Corol. \ref{thm:image-cht}). \hfill $\blacksquare$
\end{example}
The main issue is that various generative models are capable of producing the same observational image distribution, yet they can yield qualitatively distinct counterfactual images. Broadly speaking, there is nothing in the observational distribution that indicates how an image would change under a hypothetical interventional scenario, so the counterfactual distribution remains undetermined by the observational one.

\subsection{Identification of Image Counterfactual Distributions with Causal Diagrams}
One of the realizations from the broader causal inference literature is that further assumptions are needed in order to perform counterfactual reasoning. 
In this section, we will leverage the causal diagram of the true underlying ASCM to discuss whether an image counterfactual distribution is uniquely computable from a combination of these assumptions and these observational distributions. 
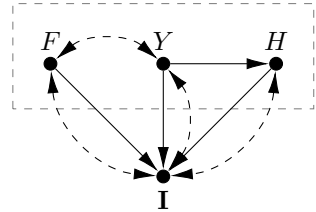
\begin{figure}[t]
	\centering
    \begin{tikzpicture}[SCM]
    \filldraw [fill=white, draw=gray,dashed] (-0.5,-0.6) rectangle (3.5,0.8);
         
        \node (X) at (0,0) [label=above:$F$, point];
        \node (Y) at (3.0,0) [label=above:$H$, point];
        \node (Z) at (1.5,0) [label=above:$Y$,point];
        \node (I) at (1.5,-1.5) [label=below:$\*I$,point];

        \path (Z) edge (Y);
        \path (X) edge (I);
		\path (Y) edge (I);
        \path (Z) edge (I);
        \path [bd] (X) edge [bend left=40] (Z);
        \path [bd] (X) edge [bend right=40] (I);
        \path [bd] (Z) edge [bend left=40] (I);
        \path [bd] (Y) edge [bend left=40] (I);
    \end{tikzpicture}
    \caption{The causal diagram of the in $\cM^*$ in \cref{ex:face-ascm}}
    \label{fig:face-ascm-g}
\vspace{-8pt}
\end{figure}

A causal diagram encodes constraints over counterfactual distributions compatible with the true and unobserved ASCM, narrowing down the hypothesis space of the proxy generator \citep[Sec. 1.4]{bareinboim:etal20}. It can be obtained from prior information about concepts in images. For instance, the qualitative understanding that getting older likely leads to gray hair suggests that there should be a direct edge from $Y$ to $H$ in \cref{ex:face-ascm}. 
Causal diagrams can be regarded as a causal inductive bias based on human knowledge.
The complete causal diagram induced by $\cM^*$ is shown in \cref{fig:face-ascm-g}; the diagram induced by $\cM^*_0$, at the generative level, is in the dashed box.
To illustrate, direct edges from $\{F, Y, H\}$ mean that these generative factors construct the image $\*I$. The bidirected edge between $F$ and $Y$ encodes that gender and age in the dataset collected are confounded. Bidirected edges between one of the generative factors $\{F, Y, H\}$ and the images imply that some unobserved generative factors can directly affect or be confounded with $\{F, Y, H\}$.

Once qualitative knowledge about the generative process is encoded in the causal model, our new goal is to infer a target image counterfactual query $P^{\cM^*}(\*I, \*I_{\*x'})$ given a causal diagram $\cG$ over $\{\*V, \*I\}$ and observational distributions $P(\*V, \*L)$.
We next define the notation of identifiability in the context of ASCMs.
\begin{definition}[Identifiability]
\label{def:id}
	Consider the true underlying ASCM $\cM^*$ defined over $\{\*V, \*I\}$ and the corresponding causal diagram $\cG$ and observational distribution $P(\*V, \*I)$. An image counterfactual query $P(\*i, \*i'_{\*x'})$ is said to be identifiable from the input $\langle P(\*V, \*I), \cG \rangle$ if $P^{\cM^{(1)}}(\*i, \*i'_{\*x'}) = P^{\cM^{(2)}}(\*i, \*i'_{\*x'})$ for every pair of ASCMs $\cM^{(1)}, \cM^{(2)} \in \Omega_{\*I}(\cG)$ s.t. $P^{\cM^{(1)}}(\*V, \*I) = P^{\cM^{(2)}}(\*V, \*I)$, where $\Omega^{\*I}$ is the space of ASCMs. The distribution $P(\*I, \*I_{\*x'})$ is said to be identifiable if $P(\*i, \*i'_{\*x'})$ is identifiable for every $\*i, \*i' \in \cX_{\*I}$.  \hfill $\blacksquare$
\end{definition}
Compared to the previous definition of identifiability used in causal inference (e.g., \citep[Ch.~3]{pearl:09}), Def. \ref{def:id} restricts 
 the space of SCMs to the space of ASCMs and considers image counterfactual queries.
The identifiability of $P(\*I, \*I_{\*x'})$ is equivalent to saying that $P(\*I, \*I_{\*x'})$ is uniquely computable given the observational distribution and the graphical constraints encoded in $\cG$. If satisfied, any proxy model $\widehat{\cM}$ that is compatible with $P(\*V, \*I)$ and $\cG$ could be used to evaluate $P^{\cM^*}(\*I, \*I_{\*x'})$ when the query is identifiable. However, the following proposition implies that even with prior causal information about $\*V$ as encoded in $\cG$, $P(\*i, \*i'_{\*x'})$ is still not identifiable. 

  \begin{restatable}[ID]{theorem}{nonidgraph}
    \label{thm:id}
    The image counterfactual distribution $P(\*I, \*I'_{\*x'})$ is not identifiable from any combination of $\langle P(\*V, \*I), \cG \rangle$.
    \hfill $\blacksquare$
\end{restatable}
This non-identifiability challenge comes from two perspectives. First, it is unknown how $U_{\*I}$ interacts with $\*V$ to produce unobserved factors $\tilde{\*U}$ while these interactions have implications for determining how the counterfactual image should look like. The next example illustrates this point. 
\begin{example}
\label{ex:face-nonid-g}
We split $\*U_{\*I}$ in Example \ref{ex:face-ascm} into $\{U_S, \*U^-_{\*I}\}$, where $U_S$ controls the smiling generative factor and $\*U^-_{\*I}$ contributes to all other unobserved generative factors. Consider two ASCMs $\cM^{(1)}$ and $\cM^{(2)}$ with the same $\cF \backslash \{f_{\*I}\}$ but different $f_{\*I}$ from $\cM^*$ defined in \cref{ex:face-ascm}:	
\begin{equation}
\begin{split}
	f_{\*I}^{(1)}(F, Y, H, U_S, \*U^-_{\*I}) &= {f}_{\*I}^{\mathrm{s}}(F, Y, H, U_S, \*U^-_{\*I}), \\
	f_{\*I}^{(2)}(F, Y, H, U_S, \*U^-_{\*I}) &= {f}_{\*I}^{\mathrm{s}}(F, Y, H, U_S \oplus Y, \*U^-_{\*I}),
\end{split}
\end{equation}
where $U_S$ is a fair coin and is independent with $\*U \backslash \{U_S\}$. 
$\*U^-_{\*I}$ can be correlated with $U_Y, U_F, U_H$. ${f}_{\*I}^{\mathrm{s}}$ is the same in both $\cM^{(1)}$ and $\cM^{(2)}$  mapping from $\{F, Y, H, S, \*U^-_{\*I}\}$ to $\*I$, where $S = U_S$ in $\cM^{(1)}$ and $S = U_S \oplus Y$ in $\cM^{(2)}$. $f_{\*I}^{\mathrm{s}}$ produces a smiling person image if and only if $S=1$. $\cM^{(1)}$ and $\cM^{(2)}$ are compatible with graphical constraints encoded in the causal diagram shown in Fig. \ref{fig:face-ascm-g}, and it is verifiable that $\cM^{(1)}$ and $\cM^{(2)}$ induce the same observational distributions.

Consider the counterfactual image query $P(\*i, \*i_{Y=0})$, where $\*i$ is a non-smiling young male and $\*i'$ is a smiling old male with gray hair. In $\cM^{(1)}$, changing $Y$ to 0 will not affect the value of $S$ while changing $Y$ to 0 will always flip the value of $S$ in $\cM^{(2)}$. This implies that $P^{\cM^{(1)}}(\*i , \*i'_{Y=0})$ is the same as $P^{\cM^{(2)}}(\*i , \*i'_{Y=0})$. \hfill $\blacksquare$
\end{example}

Second, the other perspective follows that given the observed values of a generative factor $X$ and its child $Y$, $P(y'_{x'} \mid y, x)$ is never point identifiable from the observational distribution. The next example illustrates this point.
\begin{example}
\label{ex:face-nonid-gen}
Consider an ASCM $\cM^{(3)}$ that is exactly the same as $\cM^*$ defined in \cref{ex:face-ascm} except for 
\begin{equation}
f^{(3)}_{H} \leftarrow ((\neg Y) \land U_{H_1}) \lor U_{H_2}
\end{equation}
and $P(U_{H_1}=1) = 0.25$. Then, it is verifiable that $P^{\cM^*}(\*V, \*I) = P^{\cM^{(3)}}(\*V, \*I)$ and $\cM^{(3)}$ is compatible with the graphical constraints induced by the model in \cref{fig:face-ascm-g}.

Consider the same counterfactual image query $P(\*i , \*i_{Y=0})$, and note that $H=0$ will change to $H=1$,
\begin{equation}
    P^{\cM^{(3)}}(F_{Y=0}=0, H_{Y=0}=1 \mid F=0, Y=1, H=0) = 0.25,
\end{equation}
after the intervention $do(Y=0)$ with probability $0.25$ in $\cM^{(3)}$, which is different from the same quantity induced by $\cM^*$ (\cref{eq:face-doage-hair}). This implies $P^{\cM^{(3)}}(\*i , \*i'_{Y=0})$ is not equal to $P^{\cM^*}(\*i , \*i'_{Y=0})$. \hfill $\blacksquare$
\end{example}

\section{Counterfactually consistent estimation of Image Counterfactual Distributions}
\label{sec:part-est}
We have seen so far that no image counterfactual distribution is identifiable given the causal diagram and the observational distribution alone. A question naturally arises considering this situation: 
can these non-identifiable distributions be estimated in any reasonable way? In other words, when the proxy generator (${\widehat{\cM}}$) does not induce the exact same image counterfactual distributions, what tolerance could be acceptable between $P^{\widehat{\cM}}(\*I, \*I_{\*x'})$ and the true $P^{\cM^*}(\*I, \*I_{\*x'})$? In addition, we need an estimator to guarantee the approximation of $P^{\cM^*}(\*I, \*I_{\*x'})$ be within the tolerance no matter what causal relationships among generative factors are.
To achieve this, we propose the following two directions to relax the exact estimation of query $P^{\cM^*}(\*i, \*i_{\*x'})$ while retaining causal principles and reasonable results.

\noindent (1) \textbf{Care set $\*W$.} As illustrated in \cref{sec:problem_statement}, $P^{\cM^*}(\*i, \*i_{\*x'})$ takes into account how all generative factors ($\{\*V, \tilde{\*U}\}$) in an image would change after the intervention $do(\*X=\*x)$ takes place. Still, in some practical situations, one may only be concerned about how some specific features behave after the intervention but not the whole image. 
In \cref{ex:face-ascm}, all facial features should change causally after making the person older. To illustrate, the intervention on age should preserve the gender and smiling status, and change the hair color with probability $0.4$ (\cref{eq:face-doage-hair}). However, in practice, one may only care about the gender and age (i.e., $\*W = \{F, Y\}$) after the intervention, but not whether the hair color, smiling status, and background in the image are presented the same way or not. If so, the counterfactual image can have gray hair features and smiling features with arbitrary probability. We introduce the following definition to describe the counterfactual distributions among these selected features regarding an image counterfactual query.
\begin{definition}[Feature Counterfactual Query]
\label{def:fea-ctf}
Denote $\*W$ as a set of features one cares about and $\phi$ as a function mapping from $\*I$ to $\*W$ ($\*W = \phi(\*I)$). The feature counterfactual query regarding to $P(\*i, \*i_{\*x'})$ is defined as:
\begin{equation}
\label{eq:fea-ctf}
	\hspace{-8pt} \int_{\*i^{(1)}, \*i^{(2)} \in \cX_{\*I}} \hspace{-10pt} \mathbf{1}\left [{\phi(\*i^{(1)}) = \*w, \phi(\*i^{(2)}) = \*w'}\right ]dP(\*i^{(1)}, \*i^{(2)}_{\*x'})
\end{equation}
where $\*w = \phi(\*i)$, and $\*w' = \phi(\*i')$.
We denote the feature counterfactual query  as ${\phi}(P(\*i, \*i'_{x'}))$. 
\hfill $\blacksquare$
\end{definition}
In other words, the feature counterfactual query is a push-forward measure from $P(\*i, \*i'_{x'})$ through $\phi$.
The quantity in Eq.~\ref{eq:fea-ctf} integrates over all $P(\*i^{(1)}, \*i^{(2)}_{\*x'})$ such that $\{\*i^{(1)}, \*i^{(2)}\}$ has the same cared features $\{\*w, \*w'\}$ with $\{\*i, \*i'\}$ in the target query. 
For concreteness, consider the counterfactual image query $P(\*i , \*i_{Y=0})$, where $\*i$ is a smiling young male without gray hair and $\*i'$ is a smiling old male with gray hair. Suppose the care set $\*W$ contains the features gender ($F$) and age ($Y$). The feature counterfactual query $\phi(P(\*i, \*i'_{x'}))$ calculates the probability that the original image describes a young male and the counterfactual image describes an old male after editing. Following \cref{eq:fea-ctf}, $\phi(P(\*i, \*i'_{x'}))$ sums over $P(\*i^{(1)}, \*i^{(2)}_{\*x'})$, where $\*i^{(1)}$ describes a young male, $\*i^{(2)}$ describes an old male. In addition, $\*i^{(1)}$ and $\*i^{(2)}$ can have arbitrary hair and smiling features since those are not part of $\*W$. Then, 
the feature counterfactual query induced by a proxy ASCM can be simplified using the following result. 

\begin{restatable}[]{lemma}{feactftogen}
    \label{lem:feature-gen-connection} 
    Consider the true underlying ASCM $\cM^*$ over $\{\*V, \*I\}$, and a feature set with mapping function $\phi = h^*_{\*W}$, where $h^*_{\*W}$ is the inverse function of $f^*_{\*I}$ w.r.t. $\*W$, and a  a proxy ASCM $\widehat{\cM}$ over $\{\*V, \*I\}$. if $P^{\widehat{\cM}}(\*V, \*I) = P^{\cM^*}(\*V, \*I)$, then
    \begin{equation}
        h^*_{\*W}(P^{\widehat{\cM}}(\*i, \*i'_{\*x'})) = P^{\widehat{\cM}}(\*w, \*w'_{\*x'}),
    \end{equation}
     where $\*w = h_{\*W}(\*i)$, and $\*w' = h_{\*W}(\*i')$.
    \hfill $\blacksquare$
\end{restatable}
This result suggests that if $\widehat{\cM}$ agrees on the observational distribution of $\cM^*$ and the care set $\*W$ is a subset of observed generative factors, the feature counterfactual query is equivalent to a counterfactual query $P^{\widehat{\cM}}(\*w, \*w'_{\*x'})$ over $\*W$ induced by $\widehat{\cM}_0$ at the generative level. 
We normalize $P^{\widehat{\cM}}(\*w, \*w'_{\*x'})$ as $P^{\widehat{\cM}}(\*w'_{\*x'} \mid \*w)$ following:
 \begin{equation}
     P^{\widehat{\cM}}(\*w'_{\*x'} \mid \*w) = P^{\widehat{\cM}}(\*w, \*w'_{\*x'}) / P^{\widehat{\cM}}(\*w)
 \end{equation}
 We will focus on the \textit{conditional feature counterfactual query} $P^{\widehat{\cM}}(\*w'_{\*x'} \mid \*w)$ when the proxy model satisfies $P^{\widehat{\cM}}(\*V, \*I) = P^{\cM^*}(\*V, \*I)$, which implies $P^{\widehat{\cM}}(\*w) = P^{\cM^*}(\*w)$. The following example illustrates this point.
 \begin{example}
 \label{ex:fea-ctf-query}
     Consider the counterfactual image query $P(\*i , \*i_{Y=0})$, where $\*i$ is a young male without gray hair ($F=0, Y=1, H=0$) and $\*i'$ describes an old male with gray hair ($F=0, Y=0, H=1$). Suppose the care set $\*W$ contains the feature gender ($F$) and age ($Y$) as in \cref{ex:face-ascm}. Lem. \ref{lem:feature-gen-connection} suggests the feature counterfactual query is 
     \begin{equation}
     P^{\widehat{\cM}}(F_{Y=0}=0, F = 0, Y = 1)
     \end{equation}
     whenever $\widehat{\cM}$ is compatible with $\cM^*$ w.r.t. the observational distribution. The normalized conditional feature counterfactual query is 
     \begin{equation}
     P^{\widehat{\cM}}(F_{Y=0}=0 \mid F = 0, Y = 1),
     \end{equation}
     which illustrates the probability that the gender was still male had a young male gotten older induced by the proxy model. \hfill $\blacksquare$
 \end{example}

\noindent (2) \textbf{Optimal Bounds.} 
A complementary relaxation arises from the observation that even when a query is not point identifiable, it is still possible to compute informative bounds over the target distribution from a combination of the observational data and the causal diagram \citep{manski:90, balke:pea94b, zhang:bareinboim21b}. 
These bounds serve as a natural measure of distance, or tolerance, between what is empirically obtainable from the data and the true, yet unobserved, counterfactual distribution.
This occurs because numerous ASCMs, compatible with the observed data, can generate counterfactual distributions encompassing the bound.
Any value within the optimal bound $[l, r]$ (Def. \ref{def:ctf-optimal-bound}) falls within the range of some possible ground truth, contingent on the given assumptions. As assumptions are strengthened, the bounds naturally narrow. 
Based on the above discussion, we formally define a class of counterfactual consistent estimators of the target $P(\*I, \*I_{\*x'})$. 

Based on the above discussion, we formally define the Ctf-consistent estimation of an image counterfactual query.
\begin{definition} [Ctf-Consistent Estimator w.r.t. feature set $\*W$]
	\label{def:part-est}
Consider a feature set $\*W \subseteq \*V$ and its mapping function $\phi = h^*_{\*W}$, where $h^*_{\*W}$ is the inverse function of $f^*_{\*I}$ regarding $\*W$. ${P}^{\widehat{\cM}}(\*i, \*i'_{\*x'})$ is said to be a \textit{Ctf-consistent estimator} of  ${P}^{{\cM}^*}(\*i, \*i'_{\*x'})$ w.r.t. $\*W$ if \\
(1) the observational distributions induced by $\widehat{\cM}$ and  $\cM^*$ are the same, namely, $P^{\widehat{\cM}}(\*V, \*I) = P^{\cM^*}(\*V, \*I)$ and \\
(2) the feature counterfactual query $\phi({P}^{\widehat{\cM}}(\*w, \*w'_{\*x'}))$ is within the optimal bound of ${P}(\*w, \*w'_{\*x'})$ derived by $P(\*V)$ and $\cG$, where $\*w = h_{\*W}^*(\*i)$ and $\*w' = h_{\*W}^*(\*i')$;\\
The proxy quantity $P^{\widehat{\cM}}(\*I, \*I_{\*x'})$ is said to be a Ctf-consistent estimator of the true $P^{{\cM}^*}(\*I, \*I_{\*x'})$ w.r.t. $\*W$ if $P^{\widehat{\cM}}(\*i, \*i_{\*x'})$ is Ctf-consistent for every $\*i, \*i' \in \cX_{\*I}$.
  \hfill $\blacksquare$
\end{definition}
Notice that the feature counterfactual query $\phi({P}^{\widehat{\cM}}(\*i, \*i'_{\*x'}))$ is equivalent to $P^{\widehat{\cM}}(\*w, \*w'_{\*x'})$ here according to Lem. \ref{lem:feature-gen-connection}.
Def. \ref{def:part-est} states that if 
(1) the observational distribution induced by the proxy model is the same as the true model, and  
(2) the feature counterfactual query induced by the proxy model is within the optimal bound of ${P}(\*w, \*w'_{\*x'})$, then the corresponding image counterfactual query can be regarded as a Ctf-consistent estimation of the true image counterfactual query. 
Def. \ref{def:part-est} does not require that the proxy model $\widehat{\cM}$ induces the same counterfactual image distribution $P(\*I, \*I_{\*x'})$ but expect $\widehat{\cM}$ to be Ctf-consistent with $\cM^*$ regarding the care set $\*W$ while ignoring other observed generative factors $\*V \backslash \*W$ and $\tilde{\*U}$.
Specifically, the feature counterfactual distribution $P^{\widehat{\cM}}(\*w, \*w'_{\*x'})$ should be within the optimal bound but no restriction is imposed over the features for $\*V \backslash \*W$ and $\tilde{\*U}$. 
The next example illustrates this idea.

\begin{example}
\label{ex:part-est}
(\cref{ex:fea-ctf-query} continued). 
Def. \ref{def:part-est} suggests the conditional feature counterfactual query $Q = P^{\widehat{\cM}}(F_{Y=0}=0 \mid F = 0, Y = 1)$ induced by the proxy model $\widehat{\cM}$ should be in the optimal bound $[r, l]$, where 
 \begin{equation}
 r = l = P^{\widehat{\cM}}(F=0 \mid F = 0, Y = 1) = 1 
 \end{equation}
 since the intervention $do(Y=0)$ has no effect on $F$ in the causal diagram (\cref{fig:face-ascm-g}).This implies that the gender must remain the same after the editing. In the meantime, it does not matter whether the hair is gray ($\*V \backslash \*W$) or not and whether the person is smiling ($\tilde{\*U}$) since these features are not in the care set. 

Now suppose the user cares about gender, age, and hair color, namely, $\*W = \{F, Y, H\}$ (instead of $\{F, Y\}$). Based on Def. \ref{def:part-est} and Lemma \ref{lem:feature-gen-connection}, the corresponding conditional feature counterfactual query is 
\begin{equation}
\label{n-feature-ctf}
Q = P(F_{Y=0}=0, H_{Y = 0}=1 \mid F = 0, Y = 1, H = 0),
\end{equation}
and $Q$ illustrates the probability that the individual is still a male and has gray hair after getting older. This optimal bound analytically can be derived as (see \citep[Thm. 9.2.12]{pearl:09}):
\begin{equation}
    \begin{split}
        &\hspace{-10pt}l = \max \{ 0, 1 - \frac{P(H = 0 \mid F=0, Y = 0)}{P(H=0 \mid F = 0,  Y = 1)}\} =  0.25\\
        &\hspace{-10pt}r = \min\{1, \frac{P(H=1 \mid F=0, Y=0)}{P(H=0 \mid F=0, Y=1)} \} = 0.5
    \end{split}
\end{equation}
Any $P^{\widehat{\cM}}(\*i , \*i_{Y=0})$ induced by $\widehat{\cM}$ such that $P^{\widehat{\cM}}(\*V, \*I) = P^{\cM^*}(\*V, \*I)$ and $Q^{\widehat{\cM}} \in [0.25, 0.5]$ is a Ctf-consistent estimator of $P^{\cM^*}(\*i , \*i_{Y=0})$. 
Even if $Q^{\widehat{\cM}}$ is not equal to the true feature counterfactual query $Q^{\cM^*} = 0.4$, the error is acceptable compared to the non-causal method currently used in practice.
One may change $Y$ from $1$ to $0$ and keep the other features as close as possible. With such methods, the counterfactual image will never have gray hair, thus the estimation $Q = P(F_{Y=0}=0, H_{Y = 0}=1 \mid F = 0, Y = 1, H = 0) = 0$. The causal effect of the intervention $Y=0$ on $H$ is not reflected.
 \hfill $\blacksquare$
\end{example}
 \begin{figure}[t]
\vspace{-5pt}
	\centering
    \includegraphics[scale=1.5]{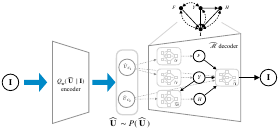}
    \caption{The ANCM network structure for \cref{ex:face-ascm}. }
    \label{fig:vae-ancm}
\vspace{-10pt}
\end{figure}

 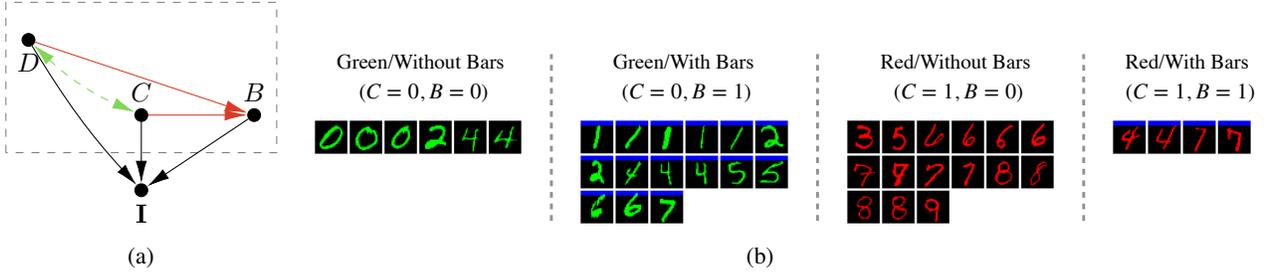
\begin{figure*}[t!]
	\begin{subfigure}[b]{0.25\linewidth}
    \centering
    \begin{tikzpicture}[SCM]
    	\filldraw [fill=white, draw=gray,dashed] (-1.8,-1.5) rectangle (1.8, 0.5);
        \node (D) at (-1.5, 0) [label=below:$D$,point];
        \node (C) at (0,-1) [label=above:$C$,point];
        \node (B) at (1.5,-1) [label=above:$B$, point];
		\node (I) at (0,-2) [label=below:$\*I$,point];
      

        \path [neg]  (D) edge (B);
        \path [neg] (C) edge (B);
        \path [pos-bd] (D) edge [bend right=10] (C);
        \path (D) edge [bend right=5.5] (I);
		\path (C) edge (I);
		\path (B) edge (I);
		                       
    \end{tikzpicture}
    \caption{}
    \label{fig:exp-mnist-backdoor}
  \end{subfigure}
  \begin{subfigure}[b]{0.7\linewidth}
    \centering
        \includegraphics[scale=1.1]{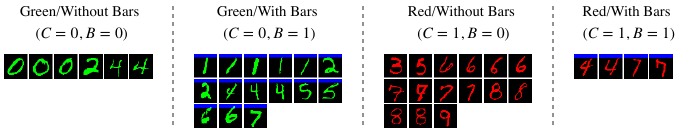}
    \caption{}
    \label{fig:backdoor-samples}
  \end{subfigure}
  \caption{The causal diagram $\cG^{\mathbf{B}}$ and samples for "Backdoor" setting. There are more red larger digits and green smaller digits; larger digits are less likely to have a bar on top; red digits are less likely to have a bar on top.}
  \vspace{-10pt}
\end{figure*}

From now on, our goal is to obtain a Ctf-consistent estimator of the non-identifiable target $P(\*I, \*I_{\*x'})$ w.r.t. the care set $\*W$. 
\begin{restatable}[Counterfactually Consistent Estimation]{theorem}{partest}
    \label{thm:part-est}
    $P^{\widehat{\cM}}(\*I, \*I'_{\*x'})$ is a Ctf-consistent estimator with respect to $\*W \subseteq \*V$ of $P^{\cM^*}(\*I, \*I'_{\*x'})$ if $\widehat{\cM} \in \Omega_{\*I}(\cG)$ and $P^{\widehat{\cM}}(\*V, \*I) = P(\*V, \*I)$.
    \hfill $\blacksquare$
\end{restatable}

The above result says that any proxy ASCM that is compatible with the diagram $\cG$ and $P(\*V, \*I)$ guarantees the estimation of the target distribution being Ctf-consistent with the true one. Specifically, in order to construct Ctf-consistent estimators, apart from fitting the generator $\widehat{\cM}$ with the given observation $P(\*V, \*I)$, it is sufficient to enforce the graphical constraints into $\widehat{\cM}$. 

\subsection{Estimating and Sampling with NCMs}
\label{sec:estimation}
We learned in the previous section that one could generate Ctf-consistent samples by fitting observational distributions to an SCM $\widehat{\cM}$ that is compatible with the given diagram (Thm. \ref{thm:part-est}). 
In this section, we develop a practical method for training $\cG$-Constrained causal deep generative models ($\cG$-NCMs) with two primary objectives: (a) to fit the observational distribution $P(\*V, \*I)$; (b) to sample images ($\*i$) and their counterfactual counterparts ($\*i'$) from them. 

Towards realizing these goals, we first acknowledge that $P^{\cM^*}(\*V, \*I)$ is typically not directly accessible in most settings, but rather its empirical counterpart $\widehat{P}^{\cM^*}(\*V, \*I) = \{\*v_k, \*i_k\}_{k=1}^n$ derived from finite datasets. 
Subsequently, we will train $\widehat{\cM}$ to match this empirical distribution $\widehat{P}^{\cM^*}(\*V, \*I)$. 
Given the substantial difference in the dimensions of variables $\*V$ (feature labels) and $\*I$ (images), we prefer to fit $P(\*I)$ and $P(\*V \mid \*I)$ separately. 
Initially, $P(\*I)$ will be learned by minimizing the data negative log-likelihood through VAEs \citep{kingma2013auto}. 
In this context, the proxy $\cG$-NCM $\widehat{\cM}$ serves as the decoder to approximate $P(\*I \mid \widehat{\*U})$ with the prior $P(\widehat{\*U})$. 
Furthermore, a separate deep neural network $Q_{\bm{\omega}}(\widehat{\*U} \mid \*I)$ is utilized to approximate the posterior $P(\widehat{\*U} \mid \*I)$, acting as the encoder, with $\bm{\omega}$ denoting the network's parameters. 
The network structure corresponding to \cref{ex:face-ascm} is illustrated in \cref{fig:vae-ancm}. The optimization objective, $L_1$, is then defined as the following evidence lower bound (ELBO) to minimize the data negative log-likelihood:
\begin{equation}
\begin{split}
       &\quad \ \ L_1(\bm{\theta}, \bm{\omega}, \widehat{P}^{\cM^*}(\*V, \*I)) \\
       &=  
       \mathbb{E}_{\widehat{P}^{\cM^*}(\*V, \*I)}\left [\mathbb{E}_{Q_{\bm{\omega}}(\widehat{\*U} \mid \*I)}[\log P^{\widehat{\cM}}(\*i \mid \widehat{\*u})] \right]\\
    &\quad - {D}_{KL}[q_{\bm{\omega}}(\widehat{\*U} \mid \*I) || P(\widehat{\*U})],
\end{split}
\end{equation}
where $\bm{\theta}$ are parameters of $\widehat{\cM}$ (see Def. \ref{def:ascm}) and ${D}_{KL}[\cdot \| \cdot]$ denotes KL divergence. To match $P(\*V \mid \*I)$, we minimize the cross-entropy loss $L_2$ of the true labels of an image sample and its predicted labels, which can be inferred through $Q_{\bm{\omega}}(\widehat{\*U} \mid \*I)$ and $\cM$ like \citep{Locatello2020DisentanglingFO, shen2022weakly}. Namely, 
\begin{equation}
\begin{split}
       &\quad \ \ L_2(\bm{\theta}, \bm{\omega}, \widehat{P}^{\cM^*}(\*V, \*I)) \\
       &=  
       \mathbb{E}_{\widehat{P}^{\cM^*}(\*V, \*I)}\left [ D_{CE}(\*V^{\widehat{\cM}}(r(\*i)), \*v)] \right]\\
\end{split}
\end{equation}
where $r(\*i)$ corresponds to the mean (vector) of $Q_{\bm{\omega}}(\widehat{\*U} \mid \*I)$ and $D_{CE}(\cdot)$ is the cross-entropy loss.
 Formally, the objective for training an NCM $\widehat{\cM}$ can be written as
\begin{equation}
\label{eq:ncm-loss}
    \hspace{-5pt} L = L_1(\bm{\theta}, \bm{\omega}, \widehat{P}^{\cM^*}(\*V, \*I)) + \lambda L_2(\bm{\theta}, \bm{\omega}, \widehat{P}^{\cM^*}(\*V, \*I))
\end{equation}
where $\lambda$ is a parameter trying to balance the likelihood $P(\*V)$ and $P(\*I | \*V)$. Specifically, a larger $\lambda$ prioritizes the fit of $P(\*V \mid \*I)$ and a smaller one prioritizes the fit of $P(\*I)$ during the training stage.
Alg. \ref{alg:ncm-learn-pv} implements more specifically the training procedure of an NCM. To illustrate, in line 1, the decoder is constructed based on the given causal diagram through Def. \ref{def:gncm}. In lines 2, all training parameters are initialized. And then in lines 3 to 6, the encoder and decoder are trained iteratively based on \cref{eq:ncm-loss}.
We refer to this approach as \textit{ANCM}. More details about network architecture and hyperparameters used throughout this work can be found in Appendix \ref{sec:experiments}.
\begin{algorithm}[tb]
    \caption{ANCM}
    \label{alg:ncm-learn-pv}
 \textbf{Input:} Data $\{\hP^{\cM^*}(\*V, \*I) = \{\*v_{k}, \*l_{k}\}_{k=1}^{n}\}$, causal diagram $\cG$, temperature $\lambda$, learning rate $\eta$, training epochs $T$.
\begin{algorithmic}[1]
   \STATE $\widehat{\cM} \gets$ $\mathrm{NCM}(\*V, \cG)$ \hfill \COMMENT{from Def.~\ref{def:gncm}}
   \STATE Initialize parameters $\bm{\theta}$ for $\widehat{\cM}$ and $\bm{\omega}$ for the inference network $Q_{\bm{\omega}}(\widehat{\*U} \mid \*I)$ \;
   \FOR{$t \gets 1$ to $T$}
    \STATE $L$$ \gets$$    L_1(\bm{\theta}, \bm{\omega}, \widehat{P}^{\cM^*}(\*V, \*I))$$ +$$ \lambda L_2(\bm{\theta}, \bm{\omega}, \widehat{P}^{\cM^*}(\*V, \*I))$
    \STATE       $\bm{\theta} \gets \bm{\theta}$$ - $$\eta \nabla L$\;
    \STATE    $\bm{\omega} \gets \bm{\omega}$$ - $$\eta \nabla L$\;
   \ENDFOR
\end{algorithmic}
\end{algorithm}

\begin{figure*}[t]
 \hfill
 \begin{subfigure}[b]{1\linewidth}
    \centering
    \includegraphics[scale=2.4]{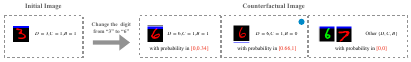}
    \caption{}
    \label{fig:backdoor-bound-1-app}
  \end{subfigure}
	\begin{subfigure}[b]{0.68\linewidth}
   \vspace{8pt}
    \centering
    \includegraphics[scale=2.1]{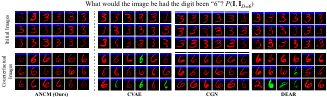}
    \caption{}
    \label{fig:backdoor-dodigit-res-app}
  \end{subfigure}
  \begin{subfigure}[b]{0.322\linewidth}
    \centering
    \includegraphics[scale=1.4]{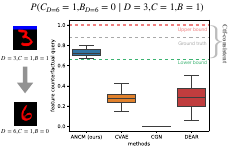}
    \caption{}
    \label{fig:backdoor-dodigit-numres-app}
  \end{subfigure}
  \caption{(a) The optimal bound of feature counterfactual queries when editing a red "3" with a bar to "6". (b) The counterfactual image generation results when editing a red "3" with a bar to "6" in the backdoor model. (c) The selected (blue circle) feature counterfactual query estimated by ANCMs and baselines.}
  \vspace{-10pt}
\end{figure*}

After training the ANCM,  we first sample $\widehat{\*u}$ from $P(\widehat{\*U})$ to generate samples of the target $P(\*I, \*I_{\*x'})$. 
The initial image sample $\widehat{\*i}$ could be derived from $\*I^{\widehat{\cM_{\*x'}}}(\widehat{\*u})$, where $\*I^{\widehat{\cM_{\*x'}}}$ is the network mapping from $\widehat{\*u}$ to $\*i$ in the decoder $\widehat{\cM}$. To edit the concept $\*X = \*x'$, the counterfactual image sample $\widehat{\*i}_{\*x'}$ could be derived through $\*I^{\widehat{\cM_{\*x'}}}(\widehat{\*u})$, where $\*I^{\widehat{\cM_{\*x'}}}$ is the network but evaluated through submodel $\widehat{\cM}_{\*x'}$ of the trained NCM.

\section{Experiments}
\label{sec:exp}
In this section, we conduct an empirical evaluation of the methods proposed in the paper, beginning with a modified Colored MNIST dataset (based on \cref{sec:cmnist-bar}) and then moving on to CelebA-HQ dataset \citep{karras2017progressive} (which describes peoples' faces) (\cref{sec:celeba-hq}). 
Further details of the model architectures are provided in Appendix \ref{sec:experiments}. 

\subsection{Colored MNIST with Bars}
\label{sec:cmnist-bar}
We first conduct experiments on the modified handwritten MNIST dataset \citep{deng2012mnist}, featuring colored digits and a horizontal blue bar in images.
\footnote{A bar in an image refers to complete rows of blue pixels.}
The observed generative factors include $\{D, C, B\}$, where $D$ denotes the digits from 0 to 9; $C$ indicates the digit color (green for $C=0$; red for $C=1$); $B$ determines whether the top of the image features a blue bar ($B=1$) or not ($B=0$). We explore two settings, named "Backdoor" (\cref{sec:backdoor-app}) and "Frontdoor" (\cref{sec:frontdoor}), each defined by unique causal relationships among the generative factors. 
This is an attractive dataset since we have full control over the generative process (SCM) and the ground truth is well-defined. 
For each task, we illustrate the concept of counterfactual editing and demonstrate that our method is capable of achieving great success in counterfactual editing tasks when compared with baselines.

 \begin{figure*}[t]
 \hfill
 \begin{subfigure}[b]{1\linewidth}
    \centering
\includegraphics[scale=2.2]{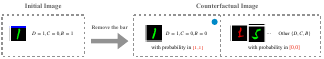}
    \caption{}
\label{fig:backdoor-bound-2}
  \end{subfigure}
	\begin{subfigure}[b]{0.68\linewidth}
    \centering
    \includegraphics[scale=2.1]{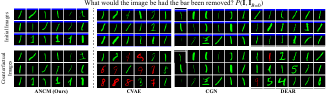}
    \caption{}
    \label{fig:backdoor-dobar-res}
  \end{subfigure}
    \hfill
  \begin{subfigure}[b]{0.32\linewidth}
    \centering
    \includegraphics[scale=1.4]{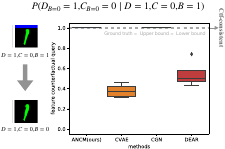}
    \caption{}
    \label{fig:backdoor-dobar-numres}
  \end{subfigure}
  \caption{(a) The optimal bound of feature counterfactual queries when removing the bar for a green "1". (b) The counterfactual image generation results when removing the bar for a green "1" in the backdoor model. (c) The selected (blue circle) feature counterfactual query estimated by ANCMs and baselines.}
    \vspace{-10pt}
\end{figure*}

\subsubsection{MNIST Backdoor Model}
\label{sec:backdoor-app}
In the Backdoor setting, the digit ($B$) and the color ($C$) are confounded with a positive correlation, but they do not directly affect each other. There are more red/larger ($\geq 5$) digits and green/smaller ($< 5$) digits in the dataset. The digit ($D$) has a negative effect on the existence of the bar ($B$). Larger digits are less likely to have a bar on the top. The color ($C$)  also has a negative effect on the existence of the bar ($B$). Red digits are less likely to have a bar on top. The true and unknown ASCM $\cM^{\mathbf{B}}$ is given by:
\begin{equation}
    \left \{
    \begin{aligned}
    D &\leftarrow U_D \\
    C &\leftarrow \bern(0.95 - 0.1U_D) \\
    B &\leftarrow (\mathbf{1}\left [ U_D \geq 5\right ] \oplus U_1) \lor (C \oplus U_2) \land U_3 \\
    \*I &\leftarrow f^\mathbf{B}_{\*I}(D, C, B, \*U_{\*I}),\\
    \end{aligned}
    \right .
\end{equation}
where the exogenous variables' distributions are:
\begin{equation}
    \begin{split}
    U_D &\sim \mathrm{Uniform}[0, 9] \\
    U_1 &\sim \bern(0.8)\\
    U_2 &\sim \bern(0.9) \\
    U_3 &\sim \bern(0.75) \\
    \end{split}
\end{equation}

The mechanism $f^\mathbf{B}$ maps the observed generative factors $\{D, C, B\}$ and unobserved generative factors (such as the position and thickness of the digit) produced by $\*U_{\*I}$ to the image $\*I$. \cref{fig:exp-mnist-backdoor} shows the causal diagram $\cG^{\mathbf{B}}$ induced by $\cM^{\mathbf{B}}$. 
The green edge indicates a positive effect and the red one represents a negative effect. We randomly sample 40 images from the collected samples in this setting and show them in \cref{fig:backdoor-samples}.

\noindent
\textbf{Task 1: Counterfactually Editing the Digits} \\
The first case we consider is to counterfactually edit the digit $D$. We let the cared features be the digit, color, and whether the image has a bar, namely, $\*W = \{B, C, D\}$. This implies that we do not care about how other generative factors (i.e., position, thickness) change in the counterfactual world. 
For counterfactual editing, changing $D$ should not affect $C$ while it might possibly change $B$, since $D$ is confounded with $C$ but has a direct effect on $B$ (\cref{fig:exp-mnist-backdoor}).

For instance, suppose we are editing a red "3" with a bar (an image with $\{D=3, C=1, B=1\}$) and wonder what would happen had the digit "3" been a "6". In this case, the optimal bounds of conditional feature counterfactual distribution $P(D_{D=6}, C_{D=6}, B_{D=6} \mid D=3, C = 1, B=1)$ derived from the observational distribution $P(D, C, B, \*I)$ and the causal diagram $\cG^{\mathbf{B}}$ are shown in \cref{fig:backdoor-bound-1-app}. Specifically, the probability that the counterfactual image has features $\{D=6, C=1, B=1\}$ is 
\begin{equation}
\begin{split}
P(C_{D=6}=1, B_{D=6} = 1 \mid D=6, C=1, B = 1) \in [0, 0.34].
 \end{split}
\end{equation}
Further, the probability that the counterfactual image has features $\{D=6, C=1, B=0\}$ is 
\begin{equation}
P(C_{D=6}=1, B_{D=6} = 0 \mid D=6, C=1, B = 1) \in [0.66, 1].
\end{equation}
The probability that the counterfactual image has other features (such as green "6", red "7") is zero. In other words, the counterfactual image by causal generative models can only be a red "6" with a bar ($\{D=6, C=1, B=0\}$) or a red "6" without a bar ($\{D=6, C=1, B=0\}$). Furthermore, the probability of the latter scenario where the bar disappears should be no less than $0.66$ due to the effect from $D$ to $B$. To achieve Ctf-consistency, we expect the generation process to follow these theoretical bounds.
 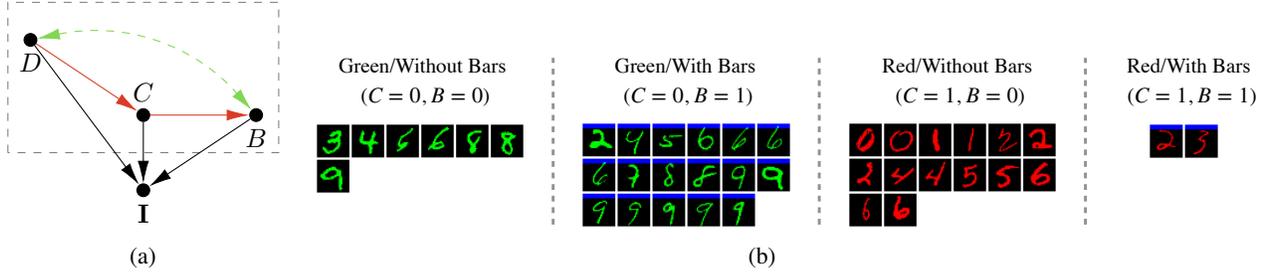
\begin{figure*}[t]
	\begin{subfigure}[b]{0.25\linewidth}
    \centering
    \begin{tikzpicture}[SCM]
    	\filldraw [fill=white, draw=gray,dashed] (-1.8,-1.5) rectangle (1.8, 0.5);
        \node (D) at (-1.5, 0) [label=below:$D$,point];
        \node (C) at (0,-1) [label=above:$C$,point];
        \node (B) at (1.5,-1) [label=below:$B$, point];
		\node (I) at (0,-2) [label=below:$\*I$,point];
      

        \path [neg] (D) edge (C);
        \path [neg] (C) edge (B);
        \path [pos-bd] (D) edge [bend left=30] (B);
        \path (D) edge (I);
		\path (C) edge (I);
		\path (B) edge (I);
		                       
    \end{tikzpicture}
    \caption{}
    \label{fig:exp-mnist-frontdoor}
  \end{subfigure}
  \begin{subfigure}[b]{0.7\linewidth}
    \centering
        \includegraphics[scale=1.1]{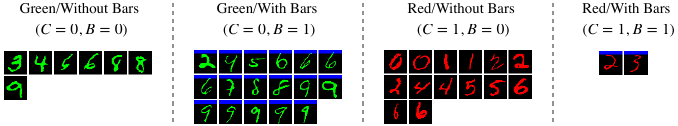}
    \caption{}
    \label{fig:frontdoor-samples}
  \end{subfigure}
  \caption{The causal diagram $\cG^{\mathbf{F}}$ and samples for "Frontdoor" setting. Bigger digits are likely to be green; red digits are less likely to have a bar on top; there are bigger digits with bars and smaller digits without bars.}
  \vspace{-10pt}
\end{figure*}

The full counterfactual image editing results are shown in \cref{fig:backdoor-dodigit-res-app}. 
After changing the digit, CVAE is likely to change the color $C$ as it uses the correlation between $D$ and $C$, while $D$ and $C$ are spuriously correlated, as discussed earlier. 
Also, the CVAE fails to capture the causal effect from $D$ to $B$ since the bar hardly disappears after the intervention $do(D=6)$. 
CGN preserves the values of both $C$ and $B$ after the intervention $D$ since it learns independent mechanisms from the generative factors to images and all generative factors are independent given the label (see more details in the Appendix \ref{sec:mnist-models-hyper}). The results suggest that the causal effect from $D$ to $B$ is not reflected in this estimation. 
DEAR follows a Markovian graph and ignores bi-direct edges in the causal diagram. Thus, DEAR fails to fit the observational distribution according to Prop.~\ref{prop:markov-expressive} and cannot preserve the color after the intervention.
ANCM preserves the original colors in counterfactual images and is likely to remove the bar, reflecting the bound value discussed above.

These results provide a qualitative suggestion that ANCM generates more realistic images that preserve the causal features found in the true generation model. Still, we would like to quantitatively understand these results, so we re-run each method 4 times and calculate the empirical probability that counterfactual images describe a red "6" without a bar after editing a red "3" with a bar to digit "6", namely $P(C_{D=6}=1, B_{D=6} = 0 \mid D=6, C=1, B = 1)$. The corresponding results are shown in \cref{fig:backdoor-dodigit-numres-app}. 
To illustrate, the gray dashed line denotes the value given by the above ASCM $\cM^{\mathbf{B}}$, which is unknown by any of the methods. The red line represents the upper bound of the feature counterfactual query and the green line represents the lower bound. We can see that queries generated by all baseline methods are not within the optimal bound. 
On the other hand, the queries generated by ANCMs are all within the optimal bound, thus they are also not far from the unknown value and are the best that can be obtained without further assumptions (over the ASCM). Both the visualization, numerical results, and theoretical results state the ANCMs are able to capture the causal effects among $\{D, C, B\}$ and produce Ctf-consistent estimators while the baselines do not.\\

\textbf{Task 2: Counterfactually Editing the Bars}\\
We now consider editing the bar $B$. We also let the care set to be $\*W = \{B, C, D\}$. 
Based on the causal diagram $\cG^{\mathbf{B}}$, we can see that changing $B$ should affect neither $D$ nor $C$. This is because $B$ is a descendant of $D$ and $C$ and not the other way around.

For concreteness, suppose we are editing an image describing a green "1" with a bar (an image with $\{D=1, C=0, B=1\}$) and wonder what would happen had the bar been removed.
In this case, the optimal bounds of conditional feature counterfactual distribution $P(D_{B=0}, C_{B=0} \mid D=1, C = 0, B=1)$ are shown in \cref{fig:backdoor-bound-2}. Specifically, the optimal bound of the probability that the counterfactual image has features $\{D=1, C=0, B=0\}$ is  
\begin{equation}
    P(D_{B=0}=1, C_{B=0}=0 \mid D=1, C=0, B=0) \in [1, 1]
\end{equation}
Further, the probability that the counterfactual image has other features (such as red "1" and green "5") is zero.
In other words, the counterfactual images must be an image describing a green "1" without a bar.

The counterfactual image editing results of ANCM and baselines are shown in \cref{fig:backdoor-dobar-res}. All methods remove bars in counterfactual images. 
Since $B$ is spuriously correlated with $D$ and $C$, CVAE and DEAR are also likely to change the color to red and to the larger digit when changing $B$ to zero. 
CGN successfully learns the independent mechanisms and changes the bars without affecting the original color and digit. We can see this branch of work (changing the intervened features but preserving others) does work for some causal relationships, but there are situations where these methods cannot work, for instance, Task 1 above (see more discussion about this in Appendix~\ref{sec:man-latent}). 
Our method ANCM also preserves these two features in counterfactual images.

These results provide a qualitative suggestion that ANCM and CGN is able to generate realistic results images that preserve the causal features found in the original model in this setting. 
In order to quantitatively understand these results, we re-run each method 4 times and calculate the empirical probability that counterfactual images describe a green "1" without bar after $do(B=0)$, namely $P(D_{B=0}=1, C_{B=0}=0 \mid D=1, C=0, B=0)$ (\cref{fig:backdoor-dobar-numres}).
The upper bound, lower bound, and the true value collapse to one line, which implies $P(D_{B=0}=1, C_{B=0}=0 \mid D=1, C=0, B=0)=1$. We can see that queries generated by CVAEs and DEAR are much smaller than the true value 1 while the queries generated by CGNs and ANCMs coincide with the true value.  Both the visualization and the numerical results suggest that CGNs and ANCMs provide Ctf-consistent estimators.

\begin{figure*}[t!]
 \hfill
 \begin{subfigure}[b]{1\linewidth}
    \centering
    \includegraphics[scale=2.0]{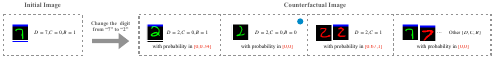}
    \caption{}
    \label{fig:frontdoor-dodigit-bound}
  \end{subfigure}
	\begin{subfigure}[b]{0.68\linewidth}
    \centering
    \includegraphics[scale=2.1]{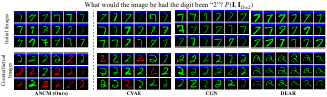}
    \caption{}
    \label{fig:frontdoor-dodigit-res}
  \end{subfigure}
    \hfill
  \begin{subfigure}[b]{0.32\linewidth}
    \centering
    \includegraphics[scale=1.4]{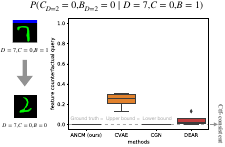}
    \caption{}
    \label{fig:frontdoor-dodigit-numres}
  \end{subfigure}
  \caption{(a) The optimal bound of feature counterfactual queries when editing a green "7" with a bar to "2". (b) The counterfactual image generation results when editing a green "7" with a bar to "2" in the frontdoor model. (c) The selected (blue circle) feature counterfactual query estimated by CVAEs and ANCMs.}
  \vspace{-10pt}
\end{figure*}

\subsubsection{MNIST Frontdoor Model}
\label{sec:frontdoor}
In the Frontdoor setting, the digit ($D$) has a negative effect on the color ($C$). Larger ($\geq 5$) digits are more likely to be green. The color $C$ has a negative effect on the existence of the bar ($B$). Red digits are less likely to have a bar on top. The digit ($D$) is confounded with the existence of the bar ($B$) with a positive correlation, but do not directly affect each other. There are larger ($\geq 5$) digits with bars and smaller ($<5$) digits without bars in the dataset. The true and unknown ASCM $\cM^{\mathbf{F}}$ is given by:
\begin{equation}
\left \{
    \begin{aligned}
    D &\leftarrow U_D \\
    C &\leftarrow \bern(0.05 + 0.1U_D) \\
    B &\leftarrow (\mathbf{1}\left [ D < 5\right ] \oplus U_2) \lor (C \oplus U_1) \land U_3 \\
    \*I &\leftarrow f^\mathbf{F}_{\*I}(D, C, B, \*U_{\*I}),\\
    \end{aligned}
\right .
\end{equation}
where the exogenous variable distributions are:
\begin{equation}
    \begin{split}
    U_D &\sim \mathrm{Uniform}[0, 9] \\
    U_1 &\sim \bern(0.8)\\
    U_2 &\sim \bern(0.9) \\
    U_3 &\sim \bern(0.7)
    \end{split}
\end{equation}
The mechanism $f^\mathbf{F}$ maps the observed generative factors $\{D, C, B\}$ and unobserved generative factors (such as the position and thickness of digits) produced by $\*U_{\*I}$ to the image $\*I$. 
\cref{fig:exp-mnist-frontdoor} shows the causal diagram $\cG^{\mathbf{F}}$ induced by $\cM^{\mathbf{F}}$. 
The green edge indicates a positive effect and the red one represents a negative effect. 
We randomly sample 40 images from the collected samples in this setting and show them in \cref{fig:frontdoor-samples}. \\

\begin{figure*}[t]
 	\begin{subfigure}[b]{1\linewidth}
    \centering
    \includegraphics[scale=2.4]{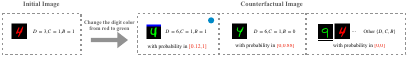}
    \caption{}
    \label{fig:front-docolor-bound}
  \end{subfigure}
	\begin{subfigure}[b]{0.68\linewidth}
    \centering
    \includegraphics[scale=2.1]{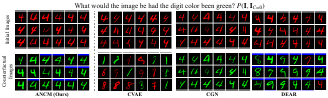}
    \caption{}
        \label{fig:front-docolor-res}
  \end{subfigure}
  \hfill
  \begin{subfigure}[b]{0.32\linewidth}
    \centering
    \includegraphics[scale=1.36]{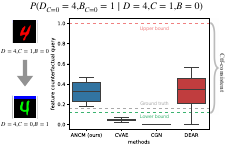}
    \label{fig:frontdoor-docolor-numres}
    \caption{}
  \end{subfigure}
  \caption{(a) The optimal bound of feature counterfactual queries when editing a red "4" without a bar to green. (b) The counterfactual image generation results when editing a red "4" with a bar to green in the frontdoor model. (c) The selected (blue circle) feature counterfactual query estimated by ANCMs and baselines.}
  \vspace{-10pt}
\end{figure*}

\noindent
\textbf{Task 3: Counterfactually Editing the Digits} \\
We first consider editing the digit $D$. 
We let the care set be $\*W = \{B, C, D\}$ similar to previous cases. 
Based on the causal diagram $\cG^{\mathbf{F}}$, changing $D$ should directly affect $C$ and is possible to change indirectly through $B$, since $D$ is the direct parent of $C$ but not a parent of $B$ (even though it's an ancestor). 
For instance, suppose we are editing a green "7" with a bar (an image with $\{D=7, C=0, B=1\}$) and wonder what would happen had the "7" been "2".
In this case, the optimal bounds of conditional feature counterfactual distribution $P(C_{D=2}, B_{D=2} \mid D=7, C = 0, B=1)$ derived from the observational distribution $P(D, C, B, \*I)$ and the causal diagram $\cG^{\mathbf{F}}$ are shown in \cref{fig:frontdoor-dodigit-bound}.
Specifically, the probability that the counterfactual image has features $\{D=2, C=0, B=1\}$ is
\begin{equation}
    P(C_{D=2}=0, B_{D=2} = 1 \mid D=7, C=0, B = 1) \in [0, 0.33].
\end{equation}
And the probability that the counterfactual image has features $\{D=2, C=0, B=0\}$ is
\begin{equation}
    P(C_{D=2}=0, B_{D=2} =0 \mid D=7, C=0, B = 1) \in [0, 0].
\end{equation}
Further, the probability that the counterfactual image has features $\{D=2, C=1\}$ is 
\begin{equation}
    P(D_{D=2}=2, C_{D=2}=1 \mid D=6, C=1, B = 1) \in [0.67, 1].
\end{equation}
And the probability that the counterfactual image has other features (such as green "7" and red "7") is zero. 
In other words, the counterfactual image by causal generative models can only be a green "2" with a bar ($\{D=6, C=1, B=0\}$) or a red "2" ($\{D=6, C=1, B=0\}, \{D=6, C=1, B=1\}$). Since $D$ can only indirectly affect $B$ through $C$, changing the digit will not influence the bar if the color remains the same. Thus, we expect that the counterfactual image generated by estimation methods would not be a green "2" without a bar.

 The counterfactual image editing results of ANCM and baselines are shown in \cref{fig:frontdoor-dodigit-res}. 
 ANCM and CVAE methods generate red digits with bars and red digits without bars, which implies they capture the effect from $D$ to $C$ and $B$. 
 However, CVAE generates counterfactual images describing green "2" without bars since $D$ and $B$ are correlated. ANCM does not change the existence of the bar when $C$ remains the same. This implies that ANCM captures the indirect effect from $D$ to $B$ as discussed above.
 CGN fails to capture the effect from $D$ to $C$ and $B$ and simply keep $C$ and $B$ the same. DEAR also fails to capture the correct causal relationships since the graph encoded in the network is incorrect and the data distribution cannot be fit.
 \begin{figure*}[!t]
 \vspace{-4pt}
    \centering
    \includegraphics[scale=1]{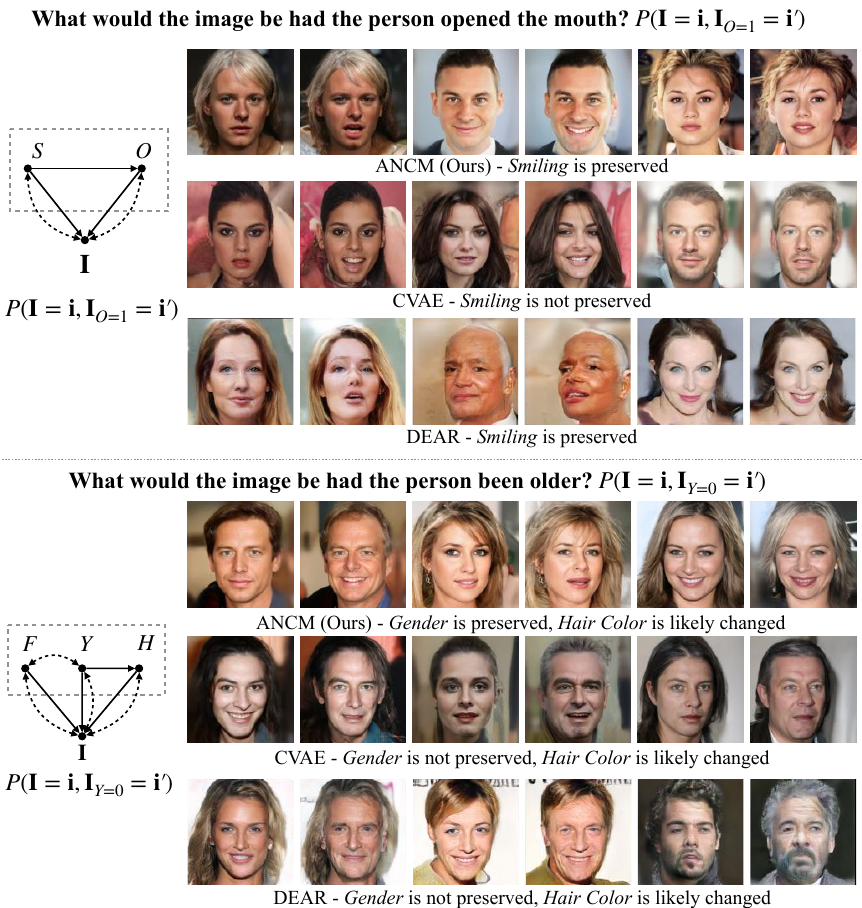}
    \caption{Editing results of the CelebaHQ Experiment. 
    }
\label{fig:celebahq-res}
\vspace{-9pt}
\end{figure*}

To quantitatively understand these results, we re-run each method 4 times and calculate the empirical probability that counterfactual images describing a green "2" without bars after editing a green "7" with a bar to digit "2", namely $P(C_{D=2}=0, B_{D=2} = 0 \mid D=7, C=0, B = 1)$. The results are shown in \cref{fig:frontdoor-dodigit-numres}. 
The upper bound, lower bound, and the feature counterfactual query generated by the true model collapse to one line, which implies $P(C_{D=2}=0, B_{D=2} = 0 \mid D=7, C=0, B = 1) = 0$. We can see that the query estimated by CVAE and DEAR are greater than 0. 
The query estimated by CGN is 0 but this is because CGNs fail to capture any (indirect or direct) effect from the digits to bars.
In contrast, the queries generated by ANCMs are exactly 0. Both the visualization and the numerical results state that ANCMs capture the full causal invariances among $\{D, C, B\}$ and produce Ctf-consistent estimators while the other methods do not. \\

\noindent
\textbf{Task 4: Counterfactually Editing the Colors} \\
We then consider editing the color $\*C$ of digits. Similarly, we let $\*W = \{B, C, D\}$. 
Based on the causal diagram $\cG^{\mathbf{F}}$, we can see that changing $C$ should not affect $D$ and is possible to change $B$ since $C$ is confounded with $D$ and has a direct effect on $B$ in \cref{fig:exp-mnist-frontdoor}.
For example, suppose we are editing a red "4" without a bar (an image with $\{D=4, C=1, B=0\}$) and wonder what would happen had the color been green. In this case, the optimal bounds of conditional feature counterfactual distribution $P(D_{C=0}, B_{C=0} \mid D=4, C = 1, B=0)$ derived from the observational distribution $P(D, C, B, \*I)$ and the causal diagram $\cG^{\mathbf{F}}$ are shown in \cref{fig:front-docolor-bound}.
Specifically, the probability that the counterfactual image has features $\{D=4, C=0, B=1\}$ is
\begin{equation}
    P(D_{C=0}=4, B_{C=0} = 1 \mid D=4, C=1, B = 0) \in [0.12, 1].
\end{equation}
Futher the probability that the counterfactual image has features $\{D=4, C=1, B=0\}$ is
\begin{equation}
    P(D_{C=0}=4, B_{C=0} = 0 \mid D=4, C=1, B = 0) \in [0, 0.88].
\end{equation}
And the probability that the counterfactual image has other features (such as green "9" and red "4") is zero. In other words, the counterfactual image by causal generative models can only be a green "4" and the bar will appear with a probability at least $0.12$ after changing the color to green.

 The counterfactual image editing results of ANCM and the baselines are shown in \cref{fig:front-docolor-res}. 
 CVAE and DEAR are also likely to change the digit since $D$ is spuriously correlated with each other and DEAR fails to fit the given distribution. 
 CGN fails to capture the direct causal effect from the color to the bar and preserves the digits and the bars as the same.
 ANCM preserves the original digit in counterfactual images after editing and the change of $B$ shows up.

These results provide a qualitative suggestion that ANCM generates more realistic results images that preserve the causal features found in the original model. 
To quantitatively understand these results, we re-run each method 4 times and calculate the empirical probability that counterfactual images describe a green "4" with a bar after editing a red "4" without a bar to green, namely $P(D_{C=0}=4, B_{C=0} = 1 \mid D=4, C=1, B = 0)$. The results are shown in \cref{fig:frontdoor-docolor-numres}. 
We can see that ANCMs present Ctf-consistent estimators while queries generated by CVAEs and CGN are nearly 0, which is not within the bound. DEARs roughly present Ctf-consistent estimators for $P(D_{C=0}=4, B_{C=0} = 1 \mid D=4, C=1, B = 0)$, but the former visualization results demonstrate DEAR cannot preserve the digit, which means it cannot provide Ctf-consistent estimators for the whole distribution.
Both the visualization and the numerical results state that ANCMs capture the causal invariances among $\{D, C, B\}$ while the baseline methods do not.

\subsection{Celeba-HQ}
\label{sec:celeba-hq}
In CelebA-HQ experiment, we consider two causal diagrams as shown in Fig. \ref{fig:celebahq-res}. In the first experiment, we consider generative factors $Smile$ $(S)$ and $Open\ Mouth$ $(O)$, and in the second experiment, we consider $Female$ $(F)$, $Young$ $(Y)$ and $Grayhair$ $(H)$.
The first target counterfactual queries are "What would the image be had the person opened the mouth?", and the second is "What would the image be had the person been older?". The feature sets are $\*W = \{S, O\}$ and $\*W = \{F, Y, H\}$ in these two settings, respectively.
We also compare ANCM (ours) against the CVAE and DEAR baselines. CGN is not compared here since the variables of CGN are restricted to $Shape, Texture$ and $Background$. Meanwhile, DiffuseVAE \citep{pandey2022diffusevae} is leveraged for ANCM and CVAE here to refine samples to high quality since VAEs often produce blurry images that lack high-frequency information \citep{dosovitskiy2016generating}. 

The empirical results are shown in Fig. \ref{fig:celebahq-res}. In the first setting, the feature set $\*W = \{S, O\}$ implies the counterfactual query is $P(S, O, S_{O=1})$, namely, "Would the person smile (or not) had the person opened the mouth?". 
The constraints induced by the ground truth model imply that changing the mouth should not affect smiling since $O$ is the direct child of $S$ and not the other way around. 
As shown \cref{fig:celebahq-res}, the smiling features are preserved after the editing by ANCM and DEAR. However, CVAE only captures the correlation between these factors, thus the non-smiling person changes to smiling after editing of mouth. On the other hand, ANCM produces higher-quality images compared to DEAR.

The second causal diagram indicates the correlations between gender and age in Example \ref{ex:face}. The dataset has more face images of young females and old males. 
More specifically,  71\% of the young people are female and 66\% of the old people are male. 
The features set $\*W = \{F, Y, H\}$ implies the counterfactual distribution is "What would the gender and the hair color of the person be had the person been older?". The causal constraints suggest that the gender of the person should be preserved and the likelihood of gray hair should increase. 
ANCM matches these causal relationships while baselines may change the original gender as shown in \cref{fig:celebahq-res}, which is of course undesirable. 

\section{Conclusions } 
We study the problem of counterfactual image generation and editing through formal causal language. Our goal in this paper is to provide guarantees that when a particular feature is edited within an image, the resulting changes in other generative factors are faithfully reflected in the counterfactual images produced. 
We formally showed that image counterfactual distributions are not identifiable from a combination of observational data and prior causal knowledge about the generating model represented as a causal diagram. 
In such non-identifiable cases, we propose new estimators (Ctf-consistent) that come accompanied with guarantees that the generated counterfactual images remain causally consistent with the true image counterfactual distribution for any causal relationship between generative factors. 
We developed an efficient algorithm to train neural causal models and sample counterfactual images. 
Finally, we demonstrate our methods are able to generate high-quality counterfactual images for synthetic images. 

Building on the machinery developed in this paper, and the understanding gained from it about image counterfactual generation, we identify some future challenges to extend these results in a broader range of practical settings. 
First, the set of features of interests, referred to as the ``care set'' in the paper  ($\*W$), guarantees the preservation of invariances and causal relations across counterfactual conditions. Although we consider settings where the set $\*W$ is labeled,  the challenge of handling unlabeled data in many practical situations remains significant.
Second, another important area for future research is enhancing the efficiency and scalability of the inferences made in this paper.

\paragraph{Impact statemnet. } 
This paper presents work whose goal is to advance the field of machine learning. There are many societal implications of our work and we hope to be beneficial, as elaborated next.  Reflecting on the broader literature, we propose the first method capable of providing formal guarantees over counterfactual image generation and editing.
The main advancement of our work lies in its emphasis on preserving the causal relationships among features, enabling sound, robust, and more realistic counterfactual generation. 
This approach differs significantly from the existing literature, which primarily focuses on reflecting the intervened features in the image. 
The critical distinction centers on what happens with the other features that were not intervened upon, and determining which features are shared or not between factual and counterfactual worlds.
Although almost never formally articulated, there are two prevalent approaches to this problem in the prior literature. 
Some works remain silent regarding the counterfactual status of the non-intervened features. This means that the neural network might leverage the correlation between features found in the factual world, leading to the various spurious results discussed earlier.  
For instance, instructing a generative AI to change a specific feature of an individual might result in a completely different person with other features, such as a different gender or race, despite they are not being causally related. 
This occurs because the neural model tends to leverage the correlation between factors found in the observational data, which is oblivious to their causal relationship. 
Other works attempt to ensure that the non-intervened features are preserved across factual and counterfactual worlds. 
However, this approach is also inadequate in settings where some of the features exert causal influence on others, and the generative AI should accordingly ascertain these relations. 
For instance, making a person older should logically lead to changes in hair color (or its amount) in both factual and counterfactual images. 

After all, we believe the results stemming from this work have broad implications for the development of the next generation of generative AI.  
First, we note that the training datasets used for large generative models are almost never balanced (see, for example, \citep{buolamwini2018gender}), which implies spurious correlations across features and the generated images. 
In practice, this often leads to more frequent, unexpected inaccuracies and biases in these models (e.g., refer to \cite{plecko2022fairness}. ) 
Understanding and accounting for the causal relationships among generative factors is fundamental for the accuracy and fairness of these models. 
Second, the lack of proper treatment of the causal invariances required for sound counterfactual reasoning translates into the impossibility of providing any sort of guarantees over what these models generate as output and their plausibility, a certainly undesirable state of affairs.  

\section*{Acknolwedgements} 
This research was supported in part by the NSF, ONR, AFOSR, DARPA, DoE, Amazon, JP Morgan, and The Alfred P. Sloan Foundation.
We thank Kevin Xia for the feedback provided in early versions of this manuscript. 
\bibliography{reference}
\bibliographystyle{icml2024}

\newpage

\appendix

\section{Proofs} \label{sec:proofs}
In this section, we provide proof of the statements in the main body of the paper.

\subsection{Proofs of Corollary \ref{thm:image-cht}}
\begin{figure}[t]
    \centering
    \includegraphics[scale=1]{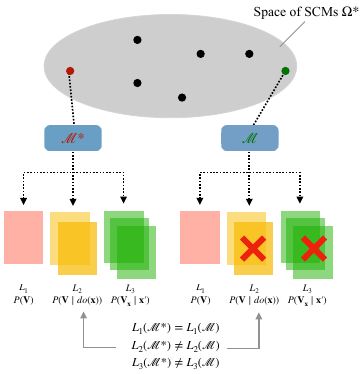}
    \caption{Causal Hierarchy Theorem (CHT).}
\label{fig:cht}
\end{figure}

First, we bring forth the formal definition of layer 1, 2, and 3 valuations, which shows how the SCM valuates observational, interventional distributions, and counterfactual distributions respectively. 
\begin{definition}[Layer 1, 2, 3 Valuation] 
\label{def:l123-semantics}
An SCM $\cM$ induces layer $\cL_3(\cM)$, a set of distributions over $\*V$, each with the form $P(\*Y_*) = P(\*Y_{1[\*x_1]}, \*Y_{2[\*x_2], \dots})$ such that 
\begin{equation}
    \label{eq:def:l3-semantics}
\begin{split}
    & P^{\cM}(\*y_{1[\*x_1]}, \*y_{2[\*x_2]}, \dots) = \\
    & \int_{\cX_{\mathbf{U}}} \mathbbm{1}\left[\*Y_{1[\*x_1]}(\*u)=\*y_1, \*Y_{2[\*x_2]}(\*u) = \*y_2, \dots \right] dP(\*u),
\end{split}
\end{equation}
where ${\*Y}_{i[\*x_i]}(\*u)$ is evaluated under 
 $\mathcal{F}_{\*x_i}\! :=\! \{f_{V_j}\! :\! V_j \in \*V \setminus \*X_i\} \cup \{f_X \leftarrow x\! :\! X \in \*X_i\}$. The specific set of distributions $P(\*Y_{\*x})$, where there is only one event, is defined as layer $L_2(\cM)$. The specific distribution $P(\*V)$, where $\*X$ is empty, is defined as layer $L_1(\cM)$.
 \hfill $\blacksquare$
\end{definition}
Then, we provide the formal Causal Hierarchy Theorem (CHT) here, which states that the layers of the hierarchy remain distinct for almost any SCM.
\begin{fact}[{Causal Hierarachy Theorem (CHT) \citep[Thm. 1]{bareinboim:etal20}}]
    \label{fact:cht}
    Let $\Omega^*$ be the set of all SCMs. We say that Layer $j$ of the causal hierarchy for SCMs collapses to Layer $i$ ($i<j$) relative to $\cM^* \in \Omega^*$ if $L_i(\cM^*) = L_i(\cM)$ implies that $L_j(\cM^*) = L_j(\cM)$ for all $\cM \in \Omega^*$. then, with respect to the Lebesgue measure over (a suitable encoding of $L_3$-equivalence classes of) SCMs, the subset in which Layer $j$ of NCMs collapses to Layer $i$ is measure zero.
\end{fact}
The CHT says that causal questions in the higher layers cannot be answered with knowledge and data restricted to lower layers. More specifically, we can almost surely find a $\cM$ for a fixed $\cM^*$ such that $L_1(\cM^*) = L_1(\cM)$ ($\cM^*$ and $\cM$ agree with observational distributions), but $L_2(\cM^*) \neq L_2(\cM), L_3(\cM^*) \neq L_3(\cM)$ (as illustrated in \cref{fig:cht}), which implies that quantities in layer 2 and 3 are not uniquely computed.

In this paper, we focus on image counterfactual distributions induced by a special class of SCMs, augmented SCMs (ASCMs) over generative factors $\*V$ and $\*I$. To prove Corollary \ref{thm:image-cht}, we aim to prove that 
it is almost surely that we can find an ASCM $\cM' \in \Omega^{\*I}$ for the true ASCM $\cM^* \in \Omega^{\*I}$ such that $L_1(\cM^*) = L_1(\cM')$, but $L_3(\cM^*) \neq L_3(\cM')$. Notice that we cannot trivially apply CHT to prove this statement since $\cM^*$ and $\cM'$ must come from the space of ASCMs $\Omega^{\*I}$. We provide the following lemma to connect image counterfactual queries induced by an ASCM $\cM$ to counterfactual quantities induced by $\cM_0$, which is a standard SCM.
\begin{restatable}[]{lem}{}
    \label{lem:ascm-scm-connection}
    Let $\cM^{(1)}$ and $\cM^{(1)}$ be two ASCMs such that $P^{\cM^{(1)}}(\*V, \*I) = P^{\cM^{(2)}}(\*V, \*I)$. Then $P^{\cM^{(1)}}(\*i, \*i'_{\*x'}) = P^{\cM^{(2)}}(\*i, \*i'_{\*x'})$ only if $P^{\cM^{(1)}_0}(\*v, \*v'_{\*x'}) = P^{\cM^{(2)}_0}(\*v, \*v'_{\*x'})$.
    \hfill $\blacksquare$
\end{restatable}

\begin{proof}
    Denote $h^{(1)}_{\*V}$ and $h^{(2)}_{\*V}$ as the function mapping from $\*I$ to $\*V$ of $\cM^{(1)}$ and $\cM^{(2)}$ respectively. We can derive $h^{(1)}_{\*V} = h^{(2)}_{\*V}$ from \cref{eq:bij} since $P^{\cM^{(1)}}(\*V \mid \*I) = P^{\cM^{(2)}}(\*V \mid \*I)$. Take the mapping function $\phi$ as $h^{(1)}_{\*V} = h^{(2)}_{\*V} = h_{\*V}$, we have
    \begin{align}
        P^{\cM^{(1)}}(\*v^*, \*v'^*_{\*x'}) = h_{\*V}(P^{\cM^{(1)}}(\*i, \*i'_{\*x'})) \\
        P^{\cM^{(2)}}(\*v^*, \*v'^*_{\*x'}) = h_{\*V}(P^{\cM^{(2)}}(\*i, \*i'_{\*x'}))
    \end{align}
    Since $P(\*v^*, \*v'^*_{\*x'})$ is the output of a function with input $P(\*i, \*i'_{\*x'})$, $P^{\cM^{(1)}}(\*i, \*i'_{\*x'}) = P^{\cM^{(2)}}(\*i, \*i'_{\*x'})$ only if $P^{\cM^{(1)}_0}(\*v, \*v'_{\*x'}) = P^{\cM^{(2)}_0}(\*v, \*v'_{\*x'})$.
\end{proof}
With Fact. \ref{fact:cht} and Lem. \ref{lem:ascm-scm-connection}, we prove Corollary \ref{thm:image-cht}. 
\imagecht*

\begin{proof}
    Let $\cM^*$ be the true underlying ASCM consisting of the generative SCM $\cM_0^*$ and $f_{\*I}^*$ and $P(\*I, \*I_{\*x'})$ be arbitrary target image counterfactual distribution. According to CHT, there always exists another generative SCM $\cM'_0$ such that $P^{\cM'_0}(\*V) = P^{\cM_0^*}(\*V)$ but $P^{\cM'_0}(\*v, \*v'_{\*x'}) \neq P^{\cM^*_0}(\*v, \*v'_{\*x'})$ for some $\*v, \*v'$. We construct an $\cM'$ over $\cM'^*_0$ as following: (1) choose $U_{\*I}$ to satisfy $P^{\cM'}(\*V, \*U_{\*I}) = P^{\cM^*}(\*V, \*U_{\*I})$; (2) let $f'_{\*I} = f_{\*I}^*$. Then $P^{\cM'}(\*V, \*I) = P^{\cM^*}(\*V, \*I)$. 
    According to Lem. \ref{lem:ascm-scm-connection}, $P^{\cM'_0}(\*v, \*v'_{\*x'}) \neq P^{\cM^*_0}(\*v, \*v'_{\*x'})$ implies that $P^{\cM_0}P(\*i, \*i'_{\*x'}) \neq P^{\cM'}P(\*i, \*i'_{\*x'})$ for any $\*i$ and $\*i'$ such that $\*v = h(\*i)$ and $\*v' = h(\*i')$. 
    Then image counterfactual distribution induced by $\cM^*$ and $\cM'$ are not the same.
    In other words, we could always construct $\cM'$ in the above way for $\cM^*$ such that $P^{\cM'}(\*V, \*I) = P^{\cM^*}(\*V, \*I)$ but $P^{\cM'}(\*I, \*I_{\*x'}) \neq P^{\cM^*}(\*I, \*I_{\*x'})$, which implies that $P(\*I, \*I_{\*x'})$ is not uniquely computable from $P(\*V, \*I)$.
\end{proof}

\subsection{Proofs of Theorem \ref{thm:id}}
In Sec. \ref{sec:problem_statement}, we explains that the mechanism $f_{\*I}$ plays two roles in the generation process of image variable $\*I$: (1) constructing unobserved generative factors $\tilde{\*U}$ from $\*U_{\*I}$ and $\*V$; (2) mixing all generative factors $\*V$ and $\tilde{\*U}$ to images pixels. We first split $f_{\*I}$ into factors constructing function $\tau$ and mixing function $f$, and write down this generation process explicitly:
\begin{equation}
\label{eq:fI-split}
\begin{split}
    \tilde{\*U} &\leftarrow \tau(\*V, \*U_{\*I}), \\
    \*I &\leftarrow \tilde{f}(\*V, \tilde{\*U}).
\end{split}
\end{equation}
We assume there exists an unobserved generative factor $\tilde{U} \in \tilde{\*U}$ such that $\tilde{U}$ is invertible from the image $\*I$, i.e. there exists function $h_{\tilde{U}}(\*I) = \tilde{U}$. In other words, a change of $\tilde{U}$ will certainly lead to a change in images. It is reasonable to assume that there exists at least such an unobserved generative factor contributing to the image, regardless of the observed generative factors $\*V$. Otherwise, the image may become excessively deterministic from labeled variable $*V$. 

With the intuition in \cref{ex:face-nonid-g}, we show the non-identifiability (Theorem \ref{thm:id}) stands from the mixing of unobserved generative factors.
\nonidgraph*
\begin{proof}
    We prove this theorem by constructing two ASCMs $\cM^{(1)}$ and $\cM^{(2)}$ that  induce the given $P(\*V, \*I)$ but differ from the image counterfactual query $P(\*I, \*I'_{\*x'})$. The true $h_{\*X}$ is known since $P(\*V \mid 
 \*I)$ is given.
    
    Consider the image counterfactual query $P(\*i, \*i'_{\*x'})$ such that $h_{\*X}(\*i) = \*x \neq h_{\*X}(\*i') = \*x'$. This query implies that the original feature $\*X = \*x$ in the original image $\*i$ is changed to $\*X = \*x'$ in the counterfactual image $\*i'$. More specifically, denote a changing variable as $X^{\Delta} \in \{ X \in \*X \mid x \neq x'\}$.
    
    Consider the unobserved generative factor $\tilde{\*U}^+$ that is invertible from $\*I$ and denote $\tilde{\*U}^{-} = \tilde{\*U} \backslash \tilde{U}^+$.
    W.L.O.G, we assume $\*V$ and $\tilde{U}$ are binary variables. 

    Suppose $P^{\cM^{(1)}}(\tilde{U}^+=0 \mid \*v) = a$ and $P^{\cM^{(1)}}(\tilde{U}^+=0 \mid \*v') = b$ where $\*v = h^{(1)}_{\*V}(\*i)$ and $\*v' = h^{(1)}_{\*V}(\*i')$. Notice that $a$ and $b$ can be arbitrary probability values since $P(\*V, \*I)$ is arbitrary. W.L.O.G, we assume $a \leq b$. According to \cref{eq:fI-split},
    $f_{\*I}^{(1)}$ in $\cM^{(1)}$ can be split as follows:
\begin{equation}
\begin{split}
    &\tilde{U}^+ \leftarrow \tau^{(1)}(\*V, \*U_{\*I}), \\
    &\tilde{\*U}^- \leftarrow \tau^-(\*V, \*U_{\*I}), \\
    &\*I \leftarrow \tilde{f}(\*V, \tilde{U}^+, \tilde{\*U}^{-}).
\end{split}
\end{equation}


Let $U^+_{\*I} \in \*U_{\*I}$ be an unobserved parent of $\tilde{\*U}^+$ and $\cX_{\tilde{\*U}^+} = \{0, 1, 2, 3, 4\}$. Becasue $\tilde{U}^+$ is a binary variable, $\tau^{(1)}$ with input $\*V = \*v$ and $\*V = \*v'$ can be always rewritten as:
\begin{equation}
\label{id-proof-tau1}
\begin{split}
        \tau^{(1)}(\*v, \*U_{\*I}) = 
    \begin{cases}
            0 & U^+_{\*I} = 0, 1;\\
            1 & U^+_{\*I} = 2, 3, 4;
    \end{cases} \\
    \tau^{(1)}(\*v', \*U_{\*I}) = 
    \begin{cases}
            0 & U^+_{\*I} = 0, 1, 2;\\
            1 & U^+_{\*I} = 3, 4;
    \end{cases} \\
\end{split}
\end{equation}
where $P(U^+_{\*I} = 0) = c, P(U^+_{\*I} = 1) = a - c, P(U^+_{\*I} = 2) = b - a, P(U^+_{\*I} = 3) = 1 - b - c, P(U^+_{\*I} = 4) = c$, where $0 < c < \min(1 - b, a)$ and $U^+_{\*I}$ is independent with $\*U_{\*I}\backslash U^+_{\*I}$. We can verify $P^{\cM^{(1)}}(\tilde{U}^+=0 \mid \*v) = a$ and $P^{\cM^{(1)}}(\tilde{U}^+=0 \mid \*v') = b$.

Then we illustrate how to construct $\cM^{(2)}$ based on $\cM^{(1)}$. First, let $\cM^{(1)}$ and $\cM^{(2)}$ equips the same generative SCM, namely $\cM^{(1)}_0 = \cM^{(2)}_0$. Second, we construct $f_{\*I}^{(2)}$ as follows:
\begin{equation}
\begin{split}
    &\tilde{U} \leftarrow \tau^{(2)}(\*V, \*U_{\*I}) \\
    &\tilde{\*U}^- \leftarrow \tau^-(\*V, \*U_{\*I}), \\
    &\*I \leftarrow \tilde{f}(\*V, \tilde{U}^+, \tilde{\*U}^-),
\end{split}
\end{equation}
where $\tau^{(1)}(\*V, \*U_{\*I}) = \tau^{(2)}(\*V, \*U_{\*I})$ when $\*V \neq \*v'$ and $\*V \neq \*v$, and
\begin{equation}
\label{id-proof-tau2}
\begin{split}
    &\tau^{(2)}(\*v, \*U_{\*I}) = 
    \begin{cases}
            0 & U^+_{\*I} = 0, 1;\\
            1 & U^+_{\*I} = 2, 3, 4;
    \end{cases} \\
    &\tau^{(2)}(\*v', \*U_{\*I}) = 
    \begin{cases}
            0 & U^+_{\*I} = 1, 2, 3;\\
            1 & U^+_{\*I} = 0, 4;
    \end{cases} \\
\end{split}
\end{equation}
In other words, $f_{\*I}^{(1)}$ and $f_{\*I}^{(2)}$ is only different from the constructing function of $\tilde{U}^+$ when $\*V = \*v'$, namely $\tau^{(1)}(\*v', \cdot) \neq \tau^{(2)}(\*v', \cdot)$.
Third, We construct $P^{(2)}(\*U)$ to satisfy $P^{\cM^{(2)}}(\*U) = P^{\cM^{(1)}}(\*U)$.

It is verifiable that 
\begin{equation}
\begin{split}
    &P^{\cM^{(2)}}(\tilde{U}^+=0 \mid \*v) = a,\\
    &P^{\cM^{(2)}}(\tilde{U}^+=0 \mid \*v') = b,\\ 
    &P^{\cM^{(2)}}(\*V) = P^{\cM^{(1)}}(\*V)
\end{split}
\end{equation}
since $P(U^+_{\*I} = 0) = P(U^+_{\*I} = 3)$ and $\cM_0^{(1)} = \cM_0^{(2)}$. 
Thus, $P^{\cM^{(1)}}(\*V, \tilde{U}^+, \tilde{\*U}^-) = P^{\cM^{(2)}}(\*V, \tilde{U}^+, \tilde{\*U}^-)$ which implies $P^{\cM^{(1)}}(\*V, \*I) = P^{\cM^{(2)}}(\*V, \*I)$. 
Also, $\cM^{(1)}$ and $\cM^{(2)}$ induces the same causal diagram since $\cM^{(1)}_0 = \cM^{(2)}_0$ and $P^{\cM^{(1)}}(\*U) = P^{\cM^{2)}}(\*U)$.

On the other hand, let $A_{x^{\Delta}, \tilde{u}^+} = \{\*i \mid \tilde{f}(\*V, \tilde{U}^+, \tilde{\*U}^-), x^{\Delta} \in  \*v, \tilde{U}^+ = \tilde{u}^+\}$. To illustrate, $A_{x^{\Delta}, \tilde{u}}$ is the range of the function $\tilde{f}(x^{\Delta}, \tilde{u}^+, \cdot)$. Notice that $A_{x^{\Delta}, 0}$, $A_{x^{\Delta}, 1}$, $A_{x'^{\Delta}, 0}$, $A_{x'^{\Delta}, 1}$ are disjoint with each other since $\tilde{U}$ is invertible from $\*I$.

Now we argue that $P^{\cM^{(1)}}(\*i, \*i'_{\*x'}) \neq P^{\cM^{(2)}}(\*i, \*i'_{\*x'})$. 
Take $\*i \in A_{x^{\Delta}, 0}$ and $\*i' \in A_{x'^{\Delta}, 1}$. Based on Def. \ref{def:l123-semantics}, $P(\*u)$ contributes to $P(\*i, \*i'_{\*x'})$ only if $\mathbf{1}[\*I(\*u) =  \*i] \land \mathbf{1}[\*I_{\*x'}(\*u) =  \*i']$. 
According to the construction of $\cM^{(2)}$, $\*I^{(1)}(\*u) = \*I^{(2)}(\*u) = \*i$ when $\*V = \*v$ and $\tilde{U}^+ = 0$ or $\tilde{U}^+ = 1$.  Given $\*u$ such that $\*I^{(1)}(\*u) = \*I^{(2)}(\*u) = \*i$, we claim that $\*I^{(1)}_{\*x'}(\*u) < \*I^{(2)}_{\*x'}(\*u)$.  
The reason is that $\tilde{U}_{\*I}^+ = 0$ or $\tilde{U}_{\*I}^+ = 1$ makes that $\tau^{1}(\*v, \*u_{\*I}) = 0$ from \cref{id-proof-tau1}, thus $\*I_{\*x'}(\*u) \notin A_{x'^{\Delta}, 1}$. Then $ P^{\cM^{(1)}}(\*i, \*i'_{\*x'}) = 0$. However, $\tilde{U}_{\*I}^+ = 0$ leads $\tau^{2}(\*v, \*u_{\*I}) = 1$ from \cref{id-proof-tau2}. Then $0 = P^{\cM^{(1)}}(\*i, \*i'_{\*x'}) < P^{\cM^{(2)}}(\*i, \*i'_{\*x'})$.
\end{proof}

The above proof constructs two ASCMs that are compatible with arbitrary $P(\*V, \*I)$ and $\cG$, but induce different image counterfactual distributions. In the construction, the non-identifiability only comes from the $f_{\*I}$ since the ASCMs have the same generative ASCMs over $\*V$. 


\subsection{Proofs of Theorem \ref{thm:part-est}}
We first prove Lem. \ref{lem:feature-gen-connection}.
\feactftogen*
\begin{proof}
$P^{\widehat{\cM}}(\*V, \*I) = P^{\cM^*}(\*V, \*I)$ implies that for any $\*i \in \cX_{\*I}, \*w \in \cX_{\*W}$, $P^{\widehat{\cM}}(\*w \mid \*i) = P^{\cM^*}(\*w \mid \*i)$. Based on \cref{eq:bij}, $h^{\widehat{\cM}}_{\*W} = h^*_{\*W}$.
Then according to Def. \ref{def:fea-ctf},
\begin{align}
&h^*_{\*W}(P^{\widehat{\cM}}(\*i, \*i'_{x'})) \nonumber\\
&=
\int_{\*i^{(1)},\*i^{(2)} \in \cX_{\*I}} \mathbf{1}\left [{h^*_{\*W}(\*i^{(1)}) = \*w, h^*_{\*W}(\*i^{(2)}) = \*w'}\right ]dP(\*i^{(1)}, \*i^{(2)}_{\*x'}) \\
&= \int_{\*u \in \cX_{\*U}} \mathbf{1}\left [{h^*_{\*W}(\*I(\*u)) = \*w, h^*_{\*W}(\*I_{\*x'}(\*u)) = \*w'}\right ]dP(\*u) \\
&\text{Def. \ref{def:l123-semantics}} \nonumber\\
&= \int_{\*u \in \cX_{\*U}} \mathbf{1}\left [\*W(\*u) = \*w, \*W_{\*x'}(\*u)) = \*w'\right ]dP(\*u) \\
&h^*_{\*W}(\*I) = h^{\widehat{\cM}}_{\*W} \nonumber\\
&= P^{\widehat{\cM}}(\*w, \*w_{\*x}) \\
&\text{Def. \ref{def:l123-semantics}} \nonumber
\end{align}
\end{proof}

Then we prove Thm. \ref{thm:part-est}.
\partest*

\begin{proof}

According to Def. \ref{def:part-est} and Lem. \ref{lem:feature-gen-connection}, ${P}^{\widehat{\cM}}(\*i, \*i'_{\*x'})$ is a causal Ctf-consistent estimation if $P^{\widehat{\cM}}(\*w, \*w'_{\*x'})$ is in the optimal bound $[l, r]$ of $P(\*w, \*w'_{\*x'})$ derived from $\cG$ and $P(\*V, \*I)$. 
We recall Def. \ref{def:ctf-optimal-bound} and write down the min/max of the optimization problem for the optimal bound in this setting.
    \begin{align}
    &\underset{\cM \in \mathbb{M}(\cG)}{\max / \min} \quad \phi({P}^{\widehat{\cM}}(\*w, \*w'_{\*x'})) \\
    & \begin{array}{r@{\quad}r@{}l@{\quad}l}
    s.t.& P^{\cM}(\*V) = P(\*V)\\
    \end{array} .
    \end{align}
It is straightforward that $P^{\widehat{\cM}}(\*w, \*w'_{\*x'})$ must be in $[l, r]$ since $\widehat{\cM}$ is in the feasible set: $\widehat{\cM} \in \mathbb{M}(\cG), P^{\widehat{\cM}}(\*V) = P(\*V)$
\end{proof}

\section{Experimental Details}
\label{sec:experiments}
This section provides details about our experiments and models. Our models are primarily written in PyTorch \citep{paszke2017automatic}, and trained with PyTorch Lightning \citep{falcon2020framework}.
 
\subsection{Colored MNIST and Bar}
\label{sec:mnist-models-hyper}

We first provide more details of the architectures of the proposed ANCM and the other three baselines: CVAE, CGN, and DEAR. 

  \textbf{ANCM (ours)}. As illustrated in Alg. \ref{alg:ncm-learn-pv} and \cref{fig:vae-ancm}, VAE-ANCM use the VAE architecture to maximize $P(\*I)$ and $P(\*V \mid \*I)$ separately, where the decoder is a $\mathcal{G}$-constrained NCM $\widehat{\cM}$ (Def. \ref{def:gncm}). 
  Each function $\widehat{f}_{V} \ \widehat{\cF}$ in $\widehat{\cM}$ is a feedforward neural network with 2 hidden layers of width 64 with layer normalization applied \citep{ba2016layer}. Each exogenous variable $\widehat{U} \in \widehat{\*U}\backslash \widehat{\*U_{\*I}}$ is a 4-dimensional standard normal distribution. The dimension of $\widehat{\*U_{\*I}}$ is set as 512. The architectures of encoder $Q_{\*w}(\tilde{\*U} \mid \*I)$ and $\widehat{f}_{\*I}$ designed based on ResNet \citep{he2016deep} and are shown in Tab. \ref{tab:cmnist-arch}. 

\textbf{CVAE \citep{sohn2015learning}}. With a latent vector $\*Z$ and data samples $\*i$, the CVAE is trained to maximize the evidence lower bound \citep{kingma2013auto} log-likelihood of the conditional image distributions, namely, $P(\*I \mid \*X)$. Specifically, the optimization objective is as follows:
 \begin{equation}
 \begin{split}
 	ELBO(\bm{\theta}, \bm{\omega}) = &\mathbb{E}_{Q_{\bm{\omega}}(\*Z \mid \*I, \*X)}[\log P_{\bm{\theta}}(\*I \mid \*Z, \*X)] \\
  & - {D}_{KL}[Q_{\bm{\omega}}(\*Z \mid \*I, \*X) \Vert P(\*Z)]
  \end{split}
 \end{equation}
 where $\*X$ is the intervened set ($D$ in this setting), $p(\*Z)$ is the guassian prior distribution of the latent vector $\*Z$, the encoder $Q_{\omega}(\*Z \mid \*I, \*X)$ is modeled as a neural network mapping from $\{\*I, \*X\}$ to $\*Z$ parametrized by $\bm{\omega}$, and the decoder $p_{\bm{\theta}}(\*I \mid \*Z, \*X)$ is modeled as a neural network mapping from $\{\*Z, \*X\}$ to $\*I$. Trained with the re-parameterization trick \citep{kingma2013auto}, the decoder is capable of generating samples from $P(\*I \mid \*X)$ ideally. 
We choose the dimension of $\*Z$ as 512, and the decoder and encoder are chosen the same as ANCM.

\begin{table*}[t!]
        \centering
        \begin{tabular}{cc}
        \hline \hline
        Encoder  & Decoder $\widehat{f}_{\*I}$ \\ \hline \hline
        $3\times3$ 32 conv, $28^2 \uparrow 32^2$, $3ch \rightarrow 32ch$ & Concat $\{\*v, \*l, \*u_{\*I}\}$, 1*1*512 fully-connected \\
        7 ResBlock, $32^2 \downarrow 16^2$, $32ch \rightarrow 64ch$ & 1 ResBlock, $1^2 \uparrow 4^2$, $512ch \rightarrow 256ch$\\ 
        7 ResBlock, $16^2 \downarrow 8^2$, $64ch \rightarrow 128ch$ & 2 ResBlock, $4^2 \downarrow 8^2$, $256ch \rightarrow 128ch$ \\ 
        3 ResBlock, $8^2 \downarrow 4^2$, $128ch \rightarrow 256ch$ & 3 ResBlock, $8^2 \downarrow 16^2$, $128ch \rightarrow 64ch$         \\ 
        3 ResBlock, $4^2 \downarrow 1^2$, $256ch \rightarrow 512ch$ & 7 ResBlock, $16^2 \downarrow 32^2$, $64ch \rightarrow 32ch$  			\\ 
        2 ResBlock, 1*1*512 fully-connected &  8 ResBlock, $3\times 3$ 3 conv. $32^2 \uparrow 28^2$, $64ch \uparrow 3ch$ 			\\ 
        \hline
        \end{tabular}
        \caption{Architectures of the networks for MNIST and Bar experiments. $3 \times 3$ represents a $3 \times 3$ convolutional kernel is adopted. $a^2 \uparrow b^2$ denotes $a \times a$ resolution is upsampled to $b \times b$. $b^2 \downarrow a^2$ denotes $b \times b$ resolution is downsampled to $a \times a$. $i\ ch \rightarrow j\ ch$ denotes the channel changes. 
        \label{tab:cmnist-arch}}
  \end{table*}

\textbf{CGN \citep{sauer2021counterfactual}}. CGN proposes to encode an SCM over variables $Shape, Texture, Background$, and $Label$ into the proxy generative model. Given the label of the image, $Shape, Texture, Background$ are independent. Formally, the mechanism of this SCM is designed as follows: 
\begin{equation}
\left \{
    \begin{aligned}
    &Label \leftarrow f_{l}(U_{l})\\
    &Shape \leftarrow \widehat{f}_s(Label, U_d) \\
    &Texture \leftarrow \widehat{f}_t(Label, U_s) \\
    &Background \leftarrow \widehat{f}_b(Label, U_b) \\
    &\*I \leftarrow \widehat{f}_{\*I}(Shape, Texture, Background),\\
    \end{aligned}
\right .
\end{equation}
where  mechanism ${f}_s, {f}_t, {f}_b$ is designed to learn the conditional distribution $P(V \mid Label)$ with prior knowledge, where $V \in \{Shape, Texture, Background\}$. The composition mechanism $\widehat{f}_{\*I}$ is not learned but defined analytically. After fitting the given observational distribution $P(Label, \*I)$, the intervention can be performed by changing the $Label$. 
In Colored MNIST and Bar experiments, the digit is regarded as $Shape$; the color is regarded as $background$ and the color is regarded as $Background$. We use the same VAE structure as ANCM to learn mechanism ${f}_s, {f}_t, {f}_b$ and the composition mechanism is designed as:
\begin{equation}
    \*I = \mathrm{Concat}(Shape \odot Texture, Background)
\end{equation}
where $\odot$ represents an elementwise multiplication operation. Theoretically, CGN learns the independent mechanism from $Shape, Texture, Background$ to the image. After performing interventions on one variable, others should be preserved in the image. 
This work can represent a branch of works that tries to change some specific features but remains others. However, they do not work on general causal relationships among generative factors (see more discussion in Appendix \ref{sec:man-latent}).

\textbf{DEAR \citep{shen2022weakly}}. DEAR is designed to learn causal representations under the supervision of annotations and Markovian causal diagrams. It encodes the given graph to the latent space by a mask adjacency matrix. DEAR also fits the $P(\*I)$ and $P(\*V \mid \*I)$ separately similar to ANCM. $P(\*I)$ is fitted with BiGAN that is capable of learning representation and generating data simultaneously \cite{donahue2016adversarial, dumoulin2016adversarially}. And $P(\*V \mid \*I)$ is fitted through a regularizer that predicts annotations from the representations. The intervention can be performed by ancestral sampling in the latent space.
In this work, we do not aim for learning the representations but only use their way to encode graphs into generative networks for comparison. When implementing DEAR, we simply ignore the bidirected edges in the given graph and encode the graph with directed edges. Theoretically, since DEAR relies on the Markovian assumption, it cannot fit the given observational distribution perfectly. This work can represent a branch of works with the Markovian assumption (see Appendix~\ref{sec:estimation-compare}). 
In theory, these methods that rely on the Markovian assumption can fail to provide Ctf-consistent estimator. See the following example:

\begin{example}
\label{ex:markov}
    Suppose one tries to use a Markovian model $\widehat{M}$ to fit observational distributions shown in Fig.~\ref{tab:face} and provide counterfactual image samples for $\cM^*$ in Example~\ref{ex:face-ascm}. The relationships between $F$ and $Y$ in $\widehat{\cM}$ should be (1) $F$ and $Y$ are independent; (2) $F \leftarrow Y$; (3) $F \rightarrow Y$;

    First, case (1) fails to fit the given observational distribution since they are strongly correlated from the data. Notice that the optimal bound of $P(f_{y'} \mid f, y)$ is $[1, 1]$, which implies $F$ is guaranteed not to change after the intervention on $Y$. 
    $\widehat{\cM}$ with case (2) fails to provide such estimation. 
    To illustrate, the optimal bound of $P(y_{f'} \mid f, y)$ is $[1, 1]$, which implies $Y$ is guaranteed not to change after the intervention on $F$.
    However, $Y$ has a direct effect on $F$ when $\widehat{\cM}$ has case (2) property, which means $Y$ is likely to change after the intervention.
    Case (3) fails to provide such estimations for the same reason. $\widehat{\cM}$ cannot provide in-bound estimation of $P(y_{f'} \mid f, y)$.
    \hfill $\blacksquare$
\end{example}

We choose DEAR since it leverages self-attention \citep{vaswani2017attention} and SAGAN \citep{zhang2019self} architectures for the discriminator and generator and then can generate high-quality images, which is important for the CelebA-HQ experiments. We choose the same network architecture and hyperparameters in the original paper for pendulum experiments. 

Encoders and decoders for ANCM, CVE, and CGN are trained with a learning rate of $10^{-4}$, and they are optimized with Adam optimizer \citep{kingma2014adam}. All training processes are performed with a batch size of 100. We choose the temperature $\lambda$ as $100$ initially and gradually decrease it during training. 

\subsection{CelebA-HQ}

\label{sec:appceleba}
    \begin{table*}[!t]
        \centering
        \begin{tabular}{cc}
        \hline \hline
        Encoder  & Decoder $\widehat{f}_{\*I}$ \\ \hline \hline
        $3\times3$ 64 conv, $128^2 \uparrow 128^2$, $3ch \rightarrow 64ch$ & Concat $\{\*v, \*l, \*u_{\*I}\}$, fully 1*1*1024 fully-connected \\
        1 ResBlock, $128^2 \downarrow 64^2$, $64ch \rightarrow 64ch$ & 1 ResBlock, $1^2 \uparrow 4^2$, $1024ch \rightarrow 512ch$\\ 
        3 ResBlock, $64^2 \downarrow 32^2$, $64ch \rightarrow 128ch$ & 2 ResBlock, $4^2 \downarrow 8^2$, $512ch \rightarrow 256ch$ \\ 
        3 ResBlock, $32^2 \downarrow 16^2$, $128ch \rightarrow 128ch$ & 2 ResBlock, $8^2 \downarrow 16^2$, $256ch \rightarrow 128ch$         \\ 
        7 ResBlock, $16^2 \downarrow 8^2$, $128ch \rightarrow 256ch$ & 7 ResBlock, $16^2 \downarrow 32^2$, $128ch \rightarrow 128ch$  			\\ 
        3 ResBlock, $8^2 \downarrow 4^2$, $256ch \rightarrow 512ch$ &  2 ResBlock, $32^2 \downarrow 64^2$, $128ch \rightarrow 64ch$			\\ 
        3 ResBlock, $4^2 \downarrow 1^2$, $512ch \rightarrow 1024ch$ & 2 ResBlock, $64^2 \downarrow 128^2$, $64ch \rightarrow 64ch$  			\\ 
        2 ResBlock, 1*1*1024 fully-connected &  1 ResBlock, $3\times 3$ 3 conv. $128^2 \uparrow 128^2$, $64ch \rightarrow 3ch$			\\ 
        \hline
        \end{tabular}
        \caption{Architectures of the networks for CelebA-HQ experiments. $3 \times 3$ represents a $3 \times 3$ convolutional kernel is adopted. $a^2 \uparrow b^2$ denotes $a \times a$ resolution is upsampled to $b \times b$. $b^2 \downarrow a^2$ denotes $b \times b$ resolution is downsampled to $a \times a$. $i\ ch \rightarrow j\ ch$ denotes the channel changes.         \label{tab:celeba-arch}}
  \end{table*}
  
\begin{figure}[!t]
  \begin{subfigure}[b]{0.45\linewidth}
    \centering
    \begin{tikzpicture}[SCM]
        \node (Z) at (-1.5,-1.5) [label=above:$S$,point];
        \node (W) at (1.5,-1.5) [label=above:$O$,point];
		\node (I) at (0,-3) [label=below:$I$,point];
      

		\path (Z) edge (I);
		\path (Z) edge (W);
		\path (W) edge (I);
		\path [bd] (Z) edge [bend right=40] (I);
		\path [bd] (W) edge [bend left=40] (I);
		                       
    \end{tikzpicture}
    \caption{}
    \label{fig:exp-face-s}
  \end{subfigure}
	\begin{subfigure}[b]{0.45\linewidth}
    \centering
    \begin{tikzpicture}[SCM]
        \node (Y) at (0,-1.5) [label=above:$Y$,point];
        \node (Z) at (-1.5,-1.5) [label=above:$F$,point];
        \node (W) at (1.5,-1.5) [label=above:$H$,point];
		\node (I) at (0,-3) [label=below:$I$,point];
      

		\path (Y) edge (W);
        \path (Y) edge (I);
		\path (Z) edge (I);
		\path (W) edge (I);
		\path [bd] (Y) edge [bend right=40] (Z);
		\path [bd] (Y) edge [bend left=40] (I);
		\path [bd] (Z) edge [bend right=40] (I);
		\path [bd] (W) edge [bend left=40] (I);
		                       
    \end{tikzpicture}
    \caption{}
      \label{fig:exp-face-y}
  \end{subfigure}
  \caption{Causal diagrams used in CelebA-HQ experiments. }
\vspace{-10pt}
\end{figure}
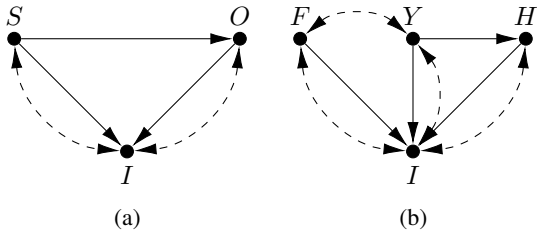

The CelebA-HQ experiment is conducted on the CelebA-HQ \citep{karras2017progressive} dataset describing human faces. The causal diagram of the ground truth is not given, however, causal diagrams shown in Fig \ref{fig:exp-face-s} and \ref{fig:exp-face-y} are used as inductive bias based on prior knowledge about human faces. To illustrate, in \emph{Smiling setting} (Fig \ref{fig:exp-face-s}), Smiling ($S$) has a positive effect on Open\_Mouth ($O$). But $S$ and $O$ can be confounded with $\*I$ by unknown generative factors $\*U_{\*I}$. 
In \emph{Age setting} (Fig \ref{fig:exp-face-y}), Young $Y$ are confounded by Female $F$ like Example \ref{ex:face} and $Y$ has a negative effect on gray hair color. Similarly, these three generative factors can be confounded with the image variable.

\begin{figure*}[!t]
    \centering
    \includegraphics[scale=0.8]{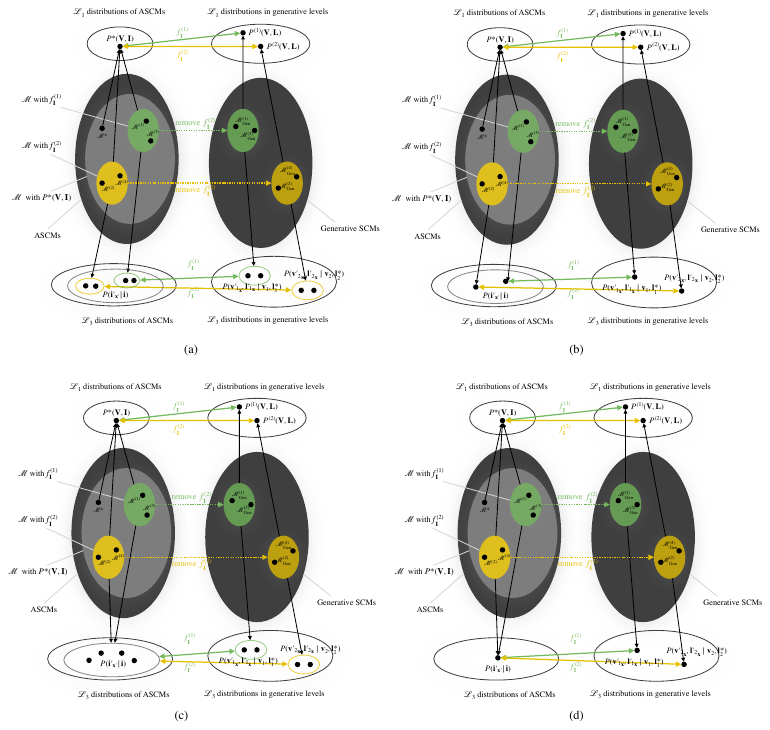}
    \caption{Internal image samples of the CelebaHQ Experiment in the Smiling setting. }
\label{fig:celebahq-ddpm}
\end{figure*}

In the Smiling setting, we are given the observed distribution $P(S, O, \*I)$ (image with labels of $S$ and $O$). We are asking the counterfactual image distribution $P(\*I, \*I_{O=1})$ (what would the image be had the person opened the Mouth?). 
We set the care set as $\{S, O\}$. In other words, the counterfactual question we ask is "Given an image, would this person smile had the person opened the mouth?". Based on the graphical constraints induced by Fig \ref{fig:exp-face-s} and the observed distributions, 
the optimal bound of $P(S=s, O=o, S_{O=o}=s)$ is $[P(S=s, O=o), P(S=s, O=o)]$; 
the $P(S=s, O=o, S_{O=o}=s')$ are $[0, 0]$, where $o \neq o'$ and $s \neq s'$. In other words, the smiling feature should be preserved no matter how $O$ changes. In the Age setting, we are given the observed distribution $P(F, Y, H, \*I)$ (image with labels of $F$, $Y$ and $H$) and we are asking the counterfactual image distribution $P(\*I, \*I_{Y=0})$. We set the care set as $\{F, Y, H\}$. In other words, the counterfactual question we ask is "Given an image, would this person have gray hair and change gender had the person become old?". Based on the graphical constraints induced by Fig \ref{fig:exp-face-y} and the observed distributions, changing $Y$ should not change $F$. And $P(F=f, Y=1, H=h, F_{Y=0}=f, H_{Y=0}=h')$ is bounded by $[r, l]$, where $r = \max(0, P(F=f, Y=1, H=h) - P(F=f, Y=1)P(H = h \mid Y = 0))$ and $l = \min(P(F=f, Y=1, H=h), P(F=f, Y=1)P(H = h' \mid Y = 0))$. Thus $P(F_{Y=0}=0, H_{Y=0}=1 \mid F=0, Y=1, H=0)$ is bounded by $[0.179, 0.1808]$ from the observational data. To illustrate, given a young male image, the probability he has gray hair is at least 0.179 but at most 0.181.

 Similarly to the last section, our causal method ANCM is compared with CVAE and DEAR in this section. We do not implement CGN since the generative variables for CGN are limited in $Shape, Texture, Background$, In this experiment, the composition function from face generative factors to images cannot be designed by hand.
 
 We leverage DiffuseVAE \citep{sanchez2022diffusion}, a two-stage method incorporating VAE and the Diffusion Probabilistic Models (DDPM) techniques, to get high-quality images. At the first stage, CVAEs are trained with association information and have the ability to edit target generative factors. ANCMs are trained with causal information (causal diagram) and have the ability to intervene on generative factors causally. However, both of them produce blurred images, $\hat{\*i}$, due to the naive VAE structure. In the second stage, these blurred image samples are refined by DDPM to high quality $\*i$. More details about the training procedure and architectures are provided in the next section. \cref{fig:celebahq-ddpm} illustrates the internal two-stage image samples for DiffuseVAE in the Smiling setting. In the VAE stages, ANCMs and CVAE generate the initial images. After performing intervention $O=1$ on each image by corresponding models, the counterfactual images are produced. Notice that in these VAE stages, image samples exhibit blurred properties. At the DDPM stage, both initial images and counterfactual images are refined to the final results.

\begin{figure*}[t]
  \vspace{0pt}
    \centering
    \includegraphics[scale=1.8]{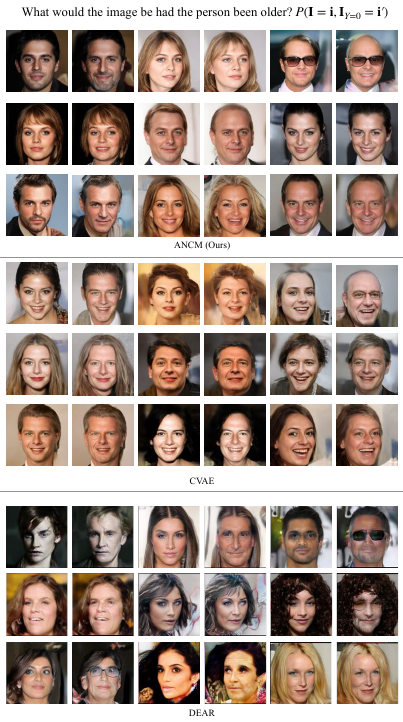}
      \vspace{-5pt}
    \caption{Additional results of the CelebaHQ Experiment in the Smiling setting. }
\label{fig:celebahq-res-s-app}
\end{figure*}

\begin{figure*}[t]
  \vspace{0pt}
    \centering
    \includegraphics[scale=1.8]{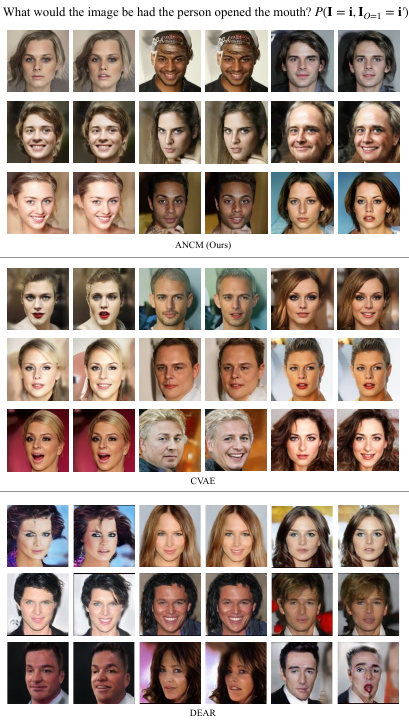}
      \vspace{-5pt}
    \caption{Additional results of the CelebaHQ Experiment in the Age setting. }
\label{fig:celebahq-res-y-app}
\end{figure*}

The experimental results are shown in Fig. \ref{fig:celebahq-res}. 
Other additional results are provided in Fig. \ref{fig:celebahq-res-s-app} (Smiling setting) and Fig. \ref{fig:celebahq-res-y-app} (Young setting). 
In the Smiling setting, all methods open the person's mouth. ANCM preserves $S$ after changing $O$. 
DEAR also successfully changes $O$ without changing $S$ since the causal diagram among $S$ and $O$ is Markovian.
However, CVAE changes the smiling features at the same time. Other features, such as background are not guaranteed to be the same since they might be confounded with $S$ and $O$. 

In the Age setting, all methods make the person's appearance older. ANCMs preserve $F$ after changing $Y$. 
However, the other two baselines change the gender features at the same time. DEAR does not work in this setting since the causal diagram is non-Markovian.
Gray hair is possible to appear after changing the Age for all three methods.

 \subsubsection{Models and Hyperparameters}
 As we discussed above, the training process has two stages: (1) the VAE stage; (2) the DDPM stage. The training procedure at the VAE stage is the same as the VAE training introduced in the Colored MNIST and Bar experiments. We choose the different architecture for encoders and $\widehat{f}_{\*I}$ as shown in Tab. \ref{tab:celeba-arch}. We set the dimension of the latent vector $\*z$ as 1024 for CVAEs. For ANCMs, exogenous variables of $\*I$ are $1024 / \max{(c, 1)}$-dimensional standard normal distributions, where $c$ is the number of bidirected arrows to $\*I$. Other exogenous variables are 10-dimensional standard normal distributions. For DEAR, the network architecture and hyperparameters are chosen as the same in \citep{shen2022weakly}.

 \begin{figure*}[!t]
    \centering
    \includegraphics[scale=2.0]{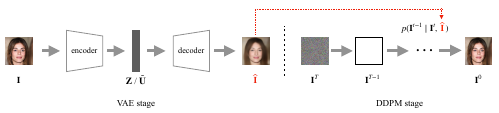}
    \caption{Two stages of DiffuseVAE. }
\label{fig:diffusionvae}
\end{figure*}

 In the second stage, the DDPM procedure is adopted to refine the decoder output at the VAE stage. DDPMs \citep{sohn2015learning, ho2020denoising} are deep generative models that consist of a forward process and reverse process with $T$ time-steps. 
 The forward process is modeled as a Markov chain that gradually perturbs $\*i^{t-1}$ (the image at step $t-1$) with gaussian noise to $\*i^{t}$ (the image at step $t$), where $\*i^0$ (image at step $0$) is the original image sample. Denoting the image variable at time step $t$ as $\*I^t$ ($\*I^0 = \*I$), the forward model can be expressed as 
 \begin{equation}
 	q(\*I^{1:T} \mid \*I_0) = \prod_{t=1}^T q(\*I^t \mid \*I^{t-1})
 \end{equation}
The reverse process is also modeled as a Markov chain that gradually denoise $\*i^{t-1}$ to $\*i_{t}$ at each time step and finally recovers the original $\*i$. Formally,
 \begin{equation}
 	p_{\varphi}(\*I^{0:T}) = p(\*I^T) \prod_{t=1}^T p_{\varphi}(\*I^{t-1} \mid \*I^t)
 \end{equation}
where  $p_{\varphi}$ implies the density is modeled by a neural network parameterized by $\varphi$ and $p(\*I^T)$ is often chosen to be an isotropic Gaussian distribution. To maximize the log-likelihood of $P(\*I)$, the following objective is minimized suggested by \cite{ho2020denoising}:
\begin{equation}
\begin{split}
    &\mathbb{E}_{q}[D_{KL}(q(\*I^{T} \mid \*I^{0}) \Vert p(\*I^{T}) ) + \\
    &\sum_{t>1}D_{KL}(q(\*I^{T-1} \mid \*I^{T}, \*I^{0}) \Vert p_{\varphi}(\*I^{T}))  \\
    &-\log p_{\varphi}(\*I^{0} \mid \*I^{1})]
\end{split}
\end{equation}
When the image $\*I$ has an associated conditioning signal $\*C$, for example, a low-resolution image \citep{ho2022cascaded, saharia2022image} or a classification label, one can maximize the log-likelihood of the conditional distribution $P(\*I \mid \*C)$ with
\begin{equation}
\label{eq:ddpm-con-obj}
\begin{split}
    	&\mathbb{E}_{q}[D_{KL}(q(\*I^{T} \mid \*I^{0}) \Vert p(\*I^{T}) ) \\
 &+ \sum_{t>1}D_{KL}(q(\*I^{T-1} \mid \*I^{T}, \*I^{0}, \*C) \Vert p_{\omega}(\*I^{T}))    \\
 &-\log p_{\omega}(\*I^{0} \mid \*I^{1}, \*C)]
\end{split}
\end{equation}

In our experiments, we leverage the reconstructions of the VAEs as the conditional signal at the second stage as suggested by DiffuseVAE. For concreteness, we freeze the encoder and decoder of the VAE trained at the first stage and train $p_{\omega}$ individually with objective Eq. \ref{eq:ddpm-con-obj}, where $\*C$ is the reconstructions of VAEs. The whole training procedure is illustrated in Fig. \ref{fig:diffusionvae}. 

All encoders and decoders are trained with a learning rate of $10^{-4}$ and are optimized with Adam optimizer \citep{kingma2014adam}. All training processes are performed with a batch size of 64. We choose the temperature $\lambda$ as $100$ and gradually decrease it. At the DDPM stage, we choose the same hyperparameters and architecture as \cite{pandey2022diffusevae}.
\section{Connection to related works} \label{sec:relatedwork}
This paper systematically establishes the counterfactual image editing tasks within the framework of causal language. 
In this section, we explicitly discuss some subtleties of extant research about image editing and counterfactual data generation. 
By integrating the causal framework proposed in this paper, we (1) interpret the empirical findings and current evaluation metrics already present in the literature; (2) show other methods may not be as general as the method discussed in this paper due to their different assumptions and problem settings. Before that, we first present the following propositions for discussion. 

\begin{proposition}
    \label{prop:single}
    Let the care set $\*W = \*X$. Any proxy model $\widehat{\cM}$ provides Ctf-consistent estimator regarding $\*W$ such that $P^{\widehat{\cM}}(\*V, \*I) = P^{{\cM}^*}(\*V, \*I)$
    \hfill $\blacksquare$
    \begin{proof}
        For any ASCM, the normalized feature counterfactual distribution 
        $P(\*W'_{\*x'} = \*w' \mid \*W = \*w ) = P(\*X'_{\*x'} = \*w' \mid \*X = \*w) = 1$ if the intervened value is $\*x'$ consistent with $\*w'$. Otherwise, this quantity is 0. This implies any ASCM provides Ctf-consistent estimators when $\*W = \*X$ from Def.~\ref{def:part-est}.
    \end{proof}
\end{proposition}

\begin{proposition}
    \label{prop:change-single}
    There exist settings in which the lower bound of normalized feature counterfactual query $P^{\cM^*}(\*w'_{\*x'} \mid \*w)$ is strictly bigger than 0 and smaller than 1 and $v \neq v'$, where $V \in \*W \backslash \*X$.
    \hfill $\blacksquare$
    \begin{proof}
        See Example.~\ref{ex:fea-ctf-query} and \ref{ex:part-est}. To illustrate, the optimal bound of the feature counterfactual query $Q$ (Eq.~\ref{n-feature-ctf}) is 0.25 and the value of $H$ differs in observation $\*w$ and counterfactual $\*w'$.
    \end{proof}
\end{proposition}

\begin{proposition}
    \label{prop:markov-expressive}
    Consider the true underlying ASCM $\cM^*$ with a semi-Markovian graph $\cG$. 
    There exist settings where any model $\widehat{\cM}$ in which the Markovianity assumption (also known as causal sufficiency) is enforced cannot satisfy (1) $P^{\widehat{\cM}}(\*V, \*I)$ =  $P^{\cM^*}(\*V, \*I)$ and (2) any feature counterfactual query is in the optimal bounds.
    \hfill $\blacksquare$
    \begin{proof}
        See the counterexample provided in Example \ref{ex:markov}.
    \end{proof}
\end{proposition}

\subsection{Counterfactual Visual Explanation} 
\label{sec:ctf-metrics}
Counterfactual visual explanation aims for the following question: 
"Given an image $\*i$ for which a vision system predicts class $C=c$, how $\*i$ could change such that the system would output a different specified class $C=c'$?". Earlier methods construct adversarial sample $\*i'$ as the counterfactual image of $\*i$. Namely, the goal is to query a permutation $\delta$ such that $\*i + \delta$ is labeled as $C=c'$ by the given predictor.
\cite{wang2020scout} search the region contributes most to the prediction in pixel levels. Other approaches map the given image to the feature space and find minimal and sufficient input features to flip the prediction \citep{dhurandhar2018explanations, goyal2019counterfactual, van2021interpretable}. However, the optimization process of search adversarial samples can push $\*i'$ far away from the data manifold of images. To enhance the realisticness of counterfactual images, VAEs \citep{goyal2019explaining,  joshi2019towards, rodriguez2021beyond}, GANs \citep{samangouei2018explaingan, khorram2022cycle} and Diffusions \citep{augustin2022diffusion} are incorporated to push counterfactual images close to the training distributions.
In the course of these studies, additional properties (as shown below) have been suggested to assess the generated counterfactual image, aside from realism \citep{verma2020counterfactual, moraffah2020causal}. It is noteworthy that these properties do not necessarily have to be satisfied simultaneously.

\textbf{Validity.} The counterfactual image should be labeled as the target $C=c'$ by the given predictor. Suppose one wants to edit an image $\*i$ describing a young person to an older look in Example \ref{ex:face}. Validity states that the pixels of counterfactual image $\*i'$ should truly constitute old features. 

\textbf{Sparsity.} The permutation $\delta$ should affect a minimal number of features. Following the previous example, sparsity encourages that counterfactual image $\*i'$ preserves features as much as possible.

\textbf{Proximity.} The counterfactual image $\*i'$ should stay as close to the initial pixel-level image. Proximity also motivates the counterfactual image not to change too much. Nevertheless, compared to sparsity, proximity focuses more on the change of each pixel. For example, the 180-degree rotation of an initial image satisfies sparsity (preserves almost all features) but does not satisfy proximity since each individual pixel varies. 

\textbf{Diversity.} The counterfactual image $\*i'$ should be as diverse as possible. To illustrate, $\*i'$ should not be unique but are motivated to be different from each other. Continue the previous example. The diversity encourages that the old version images of the original young one scan have different old extent and various other features.

Our causal framework and estimation method provide explanations for these metrics. 
First, The invertibility of $f_{\*I}$ in ASCMs supports the validity theoretically. To illustrate, the generative factor $Old=1$ causes image pixels containing old features through $f_{\*I}$, and images are predicted correctly as old class through the invertible function $h$. 
Second, the sparsity and proximity are challenged. The achievement of sparsity and proximity implies that only the intervened generative factors $\*X$ change while other generative factors keep the same. This implies that $\*X$ does not have a causal effect on other generative factors, which only happens in the causal diagram that all other generative factors are non-descendants of $\*X$. Formally, based on Prop.~\ref{prop:change-single}, sparsity and proximity should not always be satisfied. Third, the non-identifiability results (Thm. \ref{thm:id}) support the diversity. Thm. \ref{thm:id} states that ASCMs that are compatible with the same observational distribution and the same causal diagram can induce different image counterfactual distributions. Diversity can be obtained by sampling different counterfactual images from  $P(\*I_{\*x'} \mid \*I = \*i)$. 

In summary, these designed metrics are aligned with our causal framework of counterfactual image editing. However, we should notice that sparsity and proximity are limited since they ignore the causal effect from the intervened concept to other concepts in images and only work to some specific causal diagram. We propose Ctf-consistent estimators (Def. \ref{def:part-est}) for general causal relationships. For any causal relationships among generative factors, Def. \ref{def:part-est} is guaranteed to offer in-bound estimation regarding the cared feature set $\*W$.

\subsection{Manipulate latent spaces}
\label{sec:man-latent}
Generative networks, such as GANs \citep{goodfellow2020generative, karras2017progressive, brock2018large, karras2019style}, VAEs \citep{kingma2013auto, child2021very, vahdat2020nvae}, and Diffusions \citep{ho2020denoising, song2020score} fit the distribution $P(\*I)$ to learn a non-linear mapping from latent variables (generative factors) $\*Z$ to the real image variable $\*I$. 
Feeding sampled $\*z$ from the latent space, GANs, VAEs, and Diffusions are capable of producing photo-realistic image $\*i$. By manipulating the latent space values $\*z$ to $\*z'$, features in the original sample $\*i$ controlled by $\*z$ are modified by $\*z'$ and counterfactual image $\*i'$ are produced. One key point in this procedure is to find a proper way to guide the manipulation of latent vector $\*z$. 

\citet{shen2020interpreting} construct a hyperplane between class $V_1=v_1$ and class $V_1=v'_1$ (for example, old and young) and move $\*z$ along with the vertical direction to the hyperplane. This could be regarded as a process of fitting $P(\*V \mid \*I)$. The work only cares about the intervened generative factor $V_1$, namely the cared feature set $\*W$ only contains a single variable in our framework. 
Also, \citet{shen2020interpreting} proposes conditional manipulation when users care about multiple concepts in images. Conditional manipulation changes only one target and other concepts are preserved by enforcing the same distance to hyperplanes constructed by other concepts. 

Similarly, several other works perform latent vector manipulation with such linear attribute vector editing directions \citep{goetschalckx2019ganalyze, jahanian2019steerability, karras2019style}. \citet{chai2021using} train a regression model mapping from images to corresponding latent vectors and utilize it to compose the initial image with desired features. \citet{khorram2022cycle} proposes a cycle-consistent procedure to learn the transformation from the initial latent vector $\*z$ to the counterfactual latent vector $\*z'$.
Recently, text information has been leveraged into image editing as well. The image description in written text is beneficial to the encoding process and guiding the manipulation in latent space \citep{radford2021learning, avrahami2022blended, crowson2022vqgan, gal2022stylegan, kwon2022clipstyler, patashnik2021styleclip} and the natural editing instruction text can be directly used to prompt the transition from the original images to counterfactual images \citep{brooks2023instructpix2pix}. 

From Prop.~\ref{prop:single}, all these similar methods provide counterfactually consistent estimators when the case set is equal to the intervened set. However, these works ignore the effect of the intervened features on other features. Even some paper discusses the situation where other features should be preserved (such as conditional manipulation) while this only applies to restrict causal relationships among generative factors (Prop.~\ref{prop:change-single}). This point can be found in both Example~\ref{sec:part-est} and colored MNIST and bars experiments (Sec.~\ref{sec:cmnist-bar}). 
Example~\ref{sec:part-est} shows that the probability that the hair color changed to gray had the person gotten older should be at least 0.25. 
However, the approaches mentioned above cannot guarantee this counterfactual quantity and even may never change the gray hair.
On the other hand, experiments in the backdoor setting show that the probability that the bar disappeared had the digit changed should be at least 0.66. ANCMs are counterfactually consistent with the true model while these approaches (such as baseline CGN) will totally ignore this causal effect.

\subsection{Causal Representation Learning}
\label{sec:causal-represent}
Another branch of work does not focus on how to guide $\*z$ in the latent space of existing models but aims to derive an explainable latent space while training. To illustrate, these works expect that each dimension of $\*z$ represents a specific generative factor. Then manipulation can be achieved simply by modifying these corresponding dimensions of the intervened generative factors. These explainable latent vectors are called \textit{disentangled representations} for generative factors. Even though there is a lack of a formal definition of disentanglement, the key intuition is that
a disentangled representation should separate the distinct informative factors of variations in the data \cite{bengio2013representation}. Early approaches largely enforce statistical independence among latent variables $\*Z$ \citep{Higgins2017betaVAELB, Kim2018DisentanglingBF, Chen2018IsolatingSO}. However, \cite{Locatello2019ChallengingCA} argues that unsupervised disentanglement learning is impossible without additional inductive bias on the model or data. This impossible result is covered in Thm. \ref{thm:id} when the causal diagram shows independence among all generative factors. Thus, recent works shift to weakly supervised settings to overcome the non-identifiability \citep{Locatello2020WeaklySupervisedDW, Locatello2020DisentanglingFO, lachapelle2021disentanglement}. 

Despite the successful disentanglement learning in literature, they assume all generative factors are independent of each other, which only works for a special causal diagram. More recently, some methods propose to encode the structural causal model into the latent space and aim for causal disentangled representation learning \citep{yang2021causalvae, shen2022weakly, brehmer2022weakly, zhang2023identifiability, von2023nonparametric}. These works aim to learn disentangled representations of latent generative factors and even the causal diagram from interventional data. After learning the causal representations, these work demonstrate their ability to edit images causally edit images to some extent by manipulating the latent space. 

We should point out that the main contributions (learning the causal representations) of these papers are not the same (even orthogonal to) this work. The goal of this work is not to learn the disentangled latent representations behind the images but to argue how we counterfactually generate images even when the label of generative factors and the causal diagram are directly given to us. In other words, even though the causal representations are successfully learned by these works, there is still work that needs to be done for editing images. The non-identifiability result (Thm. \ref{thm:id}) implies samples obtained by manipulating the representations are hardly from the true image counterfactual distribution. There is no explanation or metrics to evaluate how these samples are causally consistent with the true model. Thm.~\ref{thm:part-est} provides such explanations, which says such methods provide in-bound estimations when the causal constraints are correctly encoded. In addition, we should also notice the SCMs that these works encoded into the neural networks are not as general as our approaches (this will be illustrated in the next section).


\subsection{Estimation Counterfactual distributions by Encoding SCMs into Neural Networks}
\label{sec:estimation-compare}
 A growing literature uses modern neural methods to estimate counterfactual queries for high-dimensional data \citep{kocaoglu2018causalgan, pawlowski2020deep, sanchez2022diffusion, sauer2021counterfactual, yang2021causalvae, shen2022weakly, brehmer2022weakly, zhang2023identifiability, von2021self}. For instance, CausalGAN \citep{kocaoglu2017causalgan} is the first work to encode the constraints induced by the causal diagram into generators of GANs. 
Implemented through GANs, flow-based and Diffusions architectures, \cite{pawlowski2020deep} and  \cite{sanchez2022diffusion} evaluates counterfactual quantities through a three-step procedure from \citet[Thm.~7.1.7]{pearl:2k}: abduction, action and prediction. 
\cite{sauer2021counterfactual} generate counterfactual images over a proxy SCM that only contains three generative factors \emph{shape}, \emph{texture}, and \emph{background}. Several perspectives limit these works in general settings.
 
\textbf{Identifiability discussion.} The identifiability of the query should be discussed before estimation, otherwise, it is unclear how the error between the estimation and the ground truth will be. Our work is the first one to talk about the identifiability of counterfactual queries when the mechanism $f_{\*I}$ is invertible (Thm. \ref{thm:id}). Even \citet{nasr2023counterfactual} talks bout the identifiability of bijective SCMs, however, the invertibility is different from ASCMs and bijective SCMs. To illustrate, bijective SCMs assume the unobserved parents $\*U_{V_j}$ are determined by both $V_j$ and parents of $V_j$ for any $V_j$. On the other hand, while ASCMs only the $f_{\*I}$ is invertible and generative factors $\*V$ are fully determined by the image variable $\*I$.

\textbf{Non-parametric SCMs with semi-Markovian diagrams.} Restrict assumptions are made for SCMs in existing works. Plenty of works assume their SCMs are Markovian, which implies the absence of unobserved confounding among generative factors. While this assumption may hold in specific settings, the same is certainly strong and does not hold in many others \citep{kocaoglu2018causalgan, pawlowski2020deep, sanchez2022diffusion, sauer2021counterfactual, shen2022weakly}. 
Prop. \ref{prop:markov-expressive} states that Markovian proxy models can fail to counterfactually estimate image counterfactual distributions in general. Also, the experiments show these Markovian proxy models (such as DEAR) fail to provide counterfactually consistent edits.
Other works focus on parametric SCMs over generative factors, such as linear mechanisms and additive noise, while we study a more general class of non-parametric models \cite{yang2021causalvae, shen2022weakly}.

\textbf{High-quality image samples.} Naive VAEs and GANs are adopted as the network structure for many existing neural causal estimation methods  \cite{kocaoglu2018causalgan, pawlowski2020deep, sauer2021counterfactual, yang2021causalvae, brehmer2022weakly}, which is not capable of generating high-quality image samples. \cite{sauer2021counterfactual} leverages pre-trained BigGAN \citep{brock2018large} to generate high-quality images that can be decomposed to shape, texture, and background. \citet{shen2022weakly} overcome the blur issue of VAE generation by self-attention techniques. \citet{ sanchez2022diffusion} and our work achieve SOTA using diffusion models.

\section{Frequently Asked Questions}
\label{app:faq}

\begin{enumerate}[label=Q\arabic*.]
    \item Is it reasonable to expect that the causal diagram is available? Why don't you learn it from the data?
    
    \textbf{Answer}. First, the assumption of the causal diagram is made out of necessity. The Image CHT result (Thm. \ref{thm:image-cht} in the main body) formally states that image counterfactual distributions are almost never recoverable from the observational data alone. 
    Causal assumptions should be made to make progress.
    The causal diagram is a well-known flexible data structure that is used throughout the literature to encode a qualitative description of the generating model, which is often much easier to obtain than the actual mechanisms of the generating SCM \citep{pearl:2k, spirtes:etal00,peters:17}. 
    Even though non-identifiability is still present given the causal diagrams (Thm. \ref{thm:id}), the underlying causal assumptions are beneficial to offer the range of some possible ground truth (in the form of optimal bounds). 
    As assumptions are strengthened, the bounds naturally narrow.  
    The goal of this paper is not to decide which set of assumptions is the best but rather to provide the toolkit for AI engines to perform the inferences once the assumptions have already been made as well as understanding the trade-off between assumptions and the guarantees provided by the method.

    Second, the true underlying causal diagrams cannot be learned only from the observational distribution in general. More specifically, there almost surely exist situations that $\cM^{(1)}$ and $\cM^{(2)}$ induce the same observational distribution but are compatible with different causal diagrams (see \citep[Sec.~1.3]{bareinboim:etal20} for details).
    With higher layer distributions (such as distributions from $L_2$), it is possible to recover an equivalence class of diagrams
    \citep{kocaoglu2017experimental, kocaoglu2019characterization, jaber2020cd, li2023causal, von2023nonparametric}. In practice, the interventional distribution is not available in many image editing tasks. As a comparison, the causal diagram itself can be more easily obtained from human knowledge in common image editing tasks. For example, getting old will lead to gray hair appearing in Example \ref{ex:face}. Our approach leverages such human inductive bias towards generative factors to obtain counterfactually consistent estimators.
    \\

    \item How could the graphical assumption be relaxed? What if the causal diagram from human knowledge is incomplete or even wrong?
    
    \textbf{Answer}. Relaxing the graphical assumption is out of the scope of this work since the goal of this paper is first to establish a causal framework of counterfactual image editing and perform counterfactually consistent estimation under general well-understood assumptions.
    Still, building on the current understanding, we believe that finding a solution to the problem when only partial information about the graph is available is an important direction for future work. One such possible approach is to utilize equivalence classes of causal diagrams, learnable from data, to perform inferences \citep{jaber2018causal,jaber2019identification, NEURIPS2022_17a9ab41, mooij2020joint, squires2020permutation}. 

    On the other hand, interestingly, even if the assumed causal diagram $\bar \cG$ is not aligned with the ground truth $\cG^*$, the generative model performs some sort of "hallucination"
    through the given input diagrams.
    More specifically, as long as the proxy model $\widehat{\cM}$ is able to fit the observational distribution with constraints induced by $\bar{\cG}$, $\widehat{\cM}$ offers counterfactually consistent estimator for the imaging world.
    For example, suppose the age will directly change human genders in a hallucinating world. Encoding edge $Gender \leftarrow Age$ into $\widehat{\cM}$ will achieve this hallucination.
    \\
    
    \item Is the non-identifiability results proved in this paper a known result in causal representation learning? 
    
    \textbf{Answer}. From a theoretical perspective, we formalize the counterfactual image editing problem with a causal framework. We list our contributions as follows and will discuss them one by one.
    
  First, we prove the non-identifiability of image counterfactual distributions from only the observational distribution (Thm.~\ref{thm:image-cht}). 
    In causal representation learning literature, there is a known result saying that disentangled causal representations are not identifiable given only the i.i.d samples from the observational distribution \citep{locatello2019challenging, lachapelle2021disentanglement, hyvarinen1999nonlinear, khemakhem2020variational}. 
    However, this result is not the same as the first contribution (a), i.e., Thm.~\ref{thm:image-cht}. 
    Specifically, the goal of our work is not to infer the latent causal representations but to query a $\cL_3$-layer distributions $P(\*I, \*I_{\*x'})$. To illustrate, the supervision information of generative factors (labels of $\*V$) is directly given to us. Even if causal representations can be successfully learned for generative factors $\*V$, the impossibility of querying the image counterfactual distributions still holds. See Appendix~\ref{sec:causal-represent} for a more detailed comparison with the causal representation learning literature. 

    Second, we show the image counterfactual distributions are not identifiable given both observational distributions (Thm.~\ref{thm:id}).
    Inferring counterfactual queries from lower-layer distributions and causal assumptions is a fundamental question in causal inference. 
    In the more traditional literature of causal inference, there are different symbolic methods for solving these problems in various settings and under different assumptions \citep{heckman:92,pearl:01,avin:etal05,shpitser:pea09,shpitser:she18,zhang2018fairness,correa:etal21}. Recently, NCMs \citep{xia:etal22} also demonstrate its ability for identification through solving an optimization procedure.  This work (Thm.~\ref{thm:id}) is the first one to show the non-identifiability of image counterfactual queries given observational distribution and the causal diagram. We refer more discussion about this identifiability result to Appemdix~.\ref{sec:estimation-compare}. 

    Third, we propose counterfactual (Ctf) consistent estimators (Def.~\ref{def:part-est}) for such non-identifiable situations and give a sufficient condition to obtain this estimator (Thm. \ref{thm:part-est}).
    This is one of the most important parts of this work. Even in non-identifiable settings, Ctf-consistent estimators offer a way to causally edit images in a reasonable way. As illustrated in Sec.~\ref{sec:part-est}, Ctf-consistent estimators ensure the feature counterfactual queries are in the optimal bound, which implies the error from the estimation to the ground truth is controlled, and this is the best we can do given the observational distribution and the causal diagram. On the other hand, a Ctf-consistent estimator could also be a metric to evaluate if other methods perform causal editing regarding the cared feature set. And this metric interprets the former metric designed for counterfactual visual explanation, as elaborated in Appendix~.\ref{sec:ctf-metrics}.
    
    From an application perspective, we design an algorithm ANCM to provide Ctf-consistent estimators for the target image counterfactual distributions and get samples from it. ANCM encodes the causal constraints induced by the given causal diagram into neural networks. Even some existing works also aim to encode SCMs into the deep generative networks, however, they are restricted in several perspectives (see Appendix~\ref{sec:estimation-compare}). Experiments in Sec.~\ref{sec:exp} and Appensix~\ref{sec:experiments} also demonstrate our method provides causally consistent editings while existing methods do not in general settings.\\
    
    \item Is the editing goal of this paper to change the intervened features in the image and keep other features the same?

    \textbf{Answer}. This is a good question and the answer is not necessarily. The situation is a bit more nuanced than that. The goal of this work is to provide causally consistent editing results with the underlying ground truth. Many existing works try to edit certain features and prevent this edit from affecting other features \citep{shen2020interpreting, goetschalckx2019ganalyze, jahanian2019steerability, karras2019style, chai2021using, khorram2022cycle}. However, this situation only works for very specific causal relationships among generative factors and there are settings where this type of editing does not provide counterfactually consistent editing (see Prop.~ \ref{prop:change-single}). For example, editing $Age$ but not changing $Gender$ is counterfactually consistent with the true model in Example \ref{ex:face-ascm}. However, if one always keeps the gray hair in the counterfactual image the same as the original image, this editing is not counterfactually consistent as illustrated in Example~\ref{ex:part-est}. This point is shown in a more practical way through the colored MNIST and bars experiment (Sec.~\ref{sec:cmnist-bar}). After changing the digit, the causal effect from Digit to Bar should be reflected. One cannot always keep the bar the same as the initial image. Further discussion is provided in Appensix.~\ref{sec:man-latent}. 
    
    In contrast, this paper proposes Ctf-consistent estimators (Def.~\ref{def:part-est}) that work for general causal relationships among generative factors.\\

    \item Why do you assume non-Markovianity? Is this more general than solutions that assume Markovianity?
    
    \textbf{Answer}.
    For clarification, the Markovianity assumption is saying that exogenous variables in SCMs $U_j$ are independent of each other and each exogenous variable affects at most one endogenous variable. In graphical terms, there are no bi-directed edges between variables in the causal diagram.
    
    We do not assume ``non-Markovianity'', but rather we do not make any assumption on Markovianity at all. 
    Typically, Markovianity is the assumption and not the other way around. Markovianity requires that there exists no unobserved confounding, but works that do not assume Markovianity do not enforce such a requirement and are therefore more general because they work in both Markovian and non-Markovian cases.  
    Markovianity is often assumed to simplify the problem setting, but it is unrealistic in some settings to expect that data on all potential confounding variables are collected in a study. Generative models that are enforced with Markovianity structure to proximate the true ASCM that induces non-Markovianity graphs are not guaranteed to provide Ctf-consistent estimators (see Prop.~\ref{prop:markov-expressive}).
    \\

\end{enumerate}

\end{document}